%% file: RGS.tex
\documentclass[acmtog,authorversion]{acmart}

\AtBeginDocument{%
  \providecommand\BibTeX{{%
    \normalfont B\kern-0.5em{\scshape i\kern-0.25em b}\kern-0.8em\TeX}}}

\copyrightyear{2024}
\acmYear{2024}
\setcopyright{acmlicensed}
\acmConference[SA Conference Papers '24]{SIGGRAPH Asia 2024 Conference Papers}{December 3--6, 2024}{Tokyo, Japan}
\acmBooktitle{SIGGRAPH Asia 2024 Conference Papers (SA Conference Papers '24), December 3--6, 2024, Tokyo, Japan}
\acmDOI{10.1145/3680528.3687576}
\acmISBN{979-8-4007-1131-2/24/12}

\usepackage{graphicx}

\begin{document}

\title{GS\textsuperscript{3}: Efficient Relighting with Triple Gaussian Splatting}

\author{Zoubin Bi}
\orcid{0009-0007-9645-8452}
\email{bzb@zju.edu.cn}

\author{Yixin Zeng}
\orcid{0009-0006-7783-2683}
\email{22221238@zju.edu.cn}

\author{Chong Zeng}
\orcid{0009-0004-6373-6848}
\email{chongzeng2000@gmail.com}

\author{Fan Pei}
\orcid{0009-0006-9302-5793}
\email{22221308@zju.edu.cn}

\author{Xiang Feng}
\orcid{0009-0003-4439-2253}
\email{xfeng.cg@zju.edu.cn}

\author{Kun Zhou}
\orcid{0000-0003-4243-6112}
\email{kunzhou@acm.org}

\author{Hongzhi Wu}
\orcid{0000-0002-4404-2275}
\email{hwu@acm.org}
\affiliation{%
  \institution{State Key Lab of CAD\&CG, Zhejiang University}
  \streetaddress{866 Yuhangtang Rd.}
  \city{Hangzhou}
  \country{China}
  \postcode{310058}
}

\newcommand{\figref}[1]{Fig.~\ref{#1}}
\newcommand{\tabref}[1]{Tab.~\ref{#1}}
\newcommand{\appref}[1]{Appendix~\ref{#1}}
\newcommand{\stepref}[1]{Step~\ref{#1}}
\newcommand{\eqnref}[1]{Eq.~\ref{#1}}
\newcommand{\algref}[1]{Algorithm~\ref{#1}}
\newcommand{\link}[1]{{\color{magenta}#1}}
\def\sec#1{Sec.~\ref{#1}}

\renewcommand{\shortauthors}{Bi et al.}

\begin{abstract}

We present a spatial and angular Gaussian based representation and a triple splatting process, for real-time, high-quality novel lighting-and-view synthesis from multi-view point-lit input images. To describe complex appearance, we employ a Lambertian plus a mixture of angular Gaussians as an effective reflectance function for each spatial Gaussian. To generate self-shadow, we splat all spatial Gaussians towards the light source to obtain shadow values, which are further refined by a small multi-layer perceptron. To compensate for other effects like global illumination, another network is trained to compute and add a per-spatial-Gaussian RGB tuple. The effectiveness of our representation is demonstrated on 30 samples with a wide variation in geometry (from solid to fluffy) and appearance (from translucent to anisotropic), as well as using different forms of input data, including rendered images of synthetic/reconstructed objects, photographs captured with a handheld camera and a flash, or from a professional lightstage. We achieve a training time of 40-70 minutes and a rendering speed of 90 fps on a single commodity GPU. Our results compare favorably with state-of-the-art techniques in terms of quality/performance.

\end{abstract}



\thanks{*Corresponding authors: Kun Zhou \& Hongzhi Wu.}
\thanks{**The first two authors contributed equally.}

\begin{teaserfigure}
    \includegraphics[width=\textwidth]{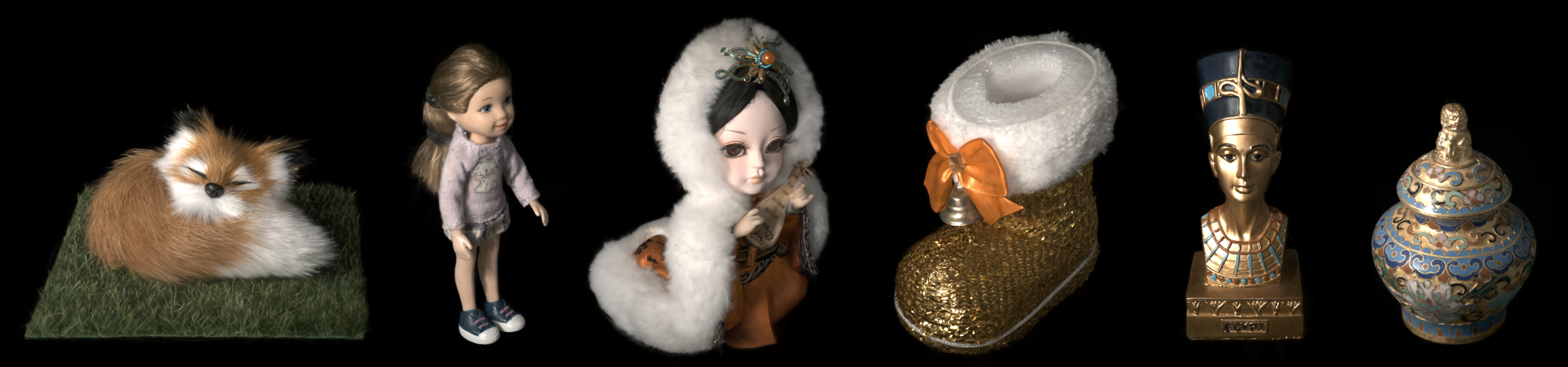}
  \caption{
  From 500-2,000 multi-view one-light-at-a-time (OLAT) input photographs, we train a representation consisting of spatial and angular Gaussians, for real-time, high-quality novel lighting-and-view synthesis. The reconstructions of a number of challenging objects with complex geometry and appearance are shown above. Please refer to the accompanying video for animated results with varying lighting and view conditions. We achieve a training time of 40-70 minutes and a rendering speed of 90 frames per second on a single commodity GPU.
  }
  \Description{Teaser.}
  \label{fig:teaser}
\end{teaserfigure}

\maketitle

\input{src/RGS-body}

\end{document}


\title{Supplemental Material for GS\textsuperscript{3}: Efficient Relighting with Triple Gaussian Splatting}

\author{Zoubin Bi}
\orcid{0009-0007-9645-8452}
\email{bzb@zju.edu.cn}

\author{Yixin Zeng}
\orcid{0009-0006-7783-2683}
\email{22221238@zju.edu.cn}

\author{Chong Zeng}
\orcid{0009-0004-6373-6848}
\email{chongzeng2000@gmail.com}

\author{Fan Pei}
\orcid{0009-0006-9302-5793}
\email{22221308@zju.edu.cn}

\author{Xiang Feng}
\orcid{0009-0003-4439-2253}
\email{xfeng.cg@zju.edu.cn}

\author{Kun Zhou}
\orcid{0000-0003-4243-6112}
\email{kunzhou@acm.org}

\author{Hongzhi Wu}
\orcid{0000-0002-4404-2275}
\email{hwu@acm.org}
\affiliation{%
  \institution{State Key Lab of CAD\&CG, Zhejiang University}
  \streetaddress{866 Yuhangtang Rd.}
  \city{Hangzhou}
  \country{China}
  \postcode{310058}
}

\renewcommand{\shortauthors}{Bi et al.}

\maketitle

\input{./src/supp_body}
\bibliographystyle{ACM-Reference-Format}
\bibliography{supplemental}

%% file: src/RGS-body.tex
\section{Introduction}
Realistically reproducing the look of a physical object at different view and lighting conditions in the virtual world has been a long-standing problem in computer graphics and computer vision. It is critical in various applications, including cultural heritage, e-commerce and visual effects.

Digital representation of shape and appearance plays a key role in this task. Traditional representations like 3D surface mesh and parametric spatially-varying bidirectional reflectance distribution function (SVBRDF) are widely used in both academia and industry~\cite{dorsey2010digital}. However, they are inherently difficult to jointly optimize with respect to input photographs, therefore often leading to suboptimal results. In the past five years, implicit representations, such as Neural Radiance Fields (NeRF)~\cite{mildenhall2020nerf}, demonstrate extraordinary ability in high-quality novel view synthesis, and even relighting~\cite{jin2023tensoir,lyu2022neural,zeng2023nrhints}. But these techniques often suffer from expensive training computation and/or slow rendering speed, limiting their applications in practice.

Recently, 3D Gaussian Splatting (GS)~\cite{kerbl3Dgaussians} gains tremendous popularity in high-quality and efficient reconstruction of Lambertian-dominant objects/scenes baked with static lighting, by essentially upgrading to a differentiable tile-based splatting method. Considerable research efforts~\cite{jiang2023gaussianshader, gao2023relightable, liang2023gsir, saito2023relightable} are made to extend GS towards novel lighting-and-view synthesis. However, high-quality relighting remains challenging, as complex appearance like anisotropic reflectance is not modeled, and the shading computation is usually confined to \emph{surface geometry} only.

In this paper, we present a novel representation based on spatial and angular Gaussians along with a triple splatting process, for real-time, high-quality novel lighting-and-view synthesis from around 500-2,000 multi-view input images, lit with one point light at a time. To describe complex appearance, we replace the spherical harmonics (SH) associated with each vanilla spatial Gaussian with a Lambertian and a mixture of angular Gaussians (a differentiable anisotropic spherical Gaussian modified from~\cite{2013asg,saito2023relightable}), essentially representing a microfacet normal distribution (i.e., 1\textsuperscript{st} splatting). To efficiently support self-shadow, we splat all spatial Gaussians toward the light source, by reusing the same high-performance pipeline as the original screen-space splatting (i.e., 2\textsuperscript{nd} splatting). To compensate for other effects like global illumination, we employ an additional multi-layer perceptron (MLP) to compute a RGB tuple for each spatial Gaussian. The above three factors are splatted to the camera and mixed to produce an image (i.e., 3\textsuperscript{rd} splatting), whose difference with a corresponding input photograph drives the optimization of our representation in a end-to-end, fully differentiable manner.

The effectiveness of our representation is demonstrated on samples with a wide variation in geometry and appearance. With  a modest increase in footprint and training/runtime computation compared with GS, we obtain high-performance and -quality synthesis results under novel lighting and view conditions. These results compare favorably with state-of-the-art techniques in terms of quality/performance. Our representation can handle a wide spectrum of input data, including rendered images of synthetic/reconstructed objects, as well as photographs captured with a smartphone and a flash, or from a professional lightstage. Our code and data are publicly available at \url{https://GSrelight.github.io/}.

\section{Related Work}
\label{sec:related work}

Below we review the most relevant work mainly in chronological order. While some existing papers require additional lighting estimation/decoupling, we would like to emphasize that this paper focuses on a general relightable representation only. We assume that the lighting is known or calibrated, and can work well with a wide spectrum of input data, from synthetic images to photographs captured with a low-end camera or a high-end lightstage. Interested readers are referred to excellent recent surveys for a broader view of the topic~\cite{tewari2022nrsota,fei20243dgssurvey,wu2024gsrecent}.

\subsection{Traditional Relighting}

While widely deployed in practice, traditional representations, such as 3D surface mesh and parametric SVBRDF which varies with location, view and lighting directions, are challenging to optimize jointly. The majority of existing work performs separate estimations of shape and appearance, the latter of which is typically represented as attributes defined on a known 3D geometry. Dense lights are used to remove adversarial effects like strong specular reflections to enable geometry reconstruction with multi-view stereo, prior to reflectance estimation~\cite{kang2019learning,tunwattanapong2013acquiring}. Zhou et al.~\shortcite{zhou2013multi}  recover a 3D shape from multi-view photometric cues, and then compute isotropic surface reflectance. Structured illumination is adopted to recover highly precise surface geometry, after which the appearance is computed~\cite{Holroyd10Acoaxial,xu2023unified}. Despite training an image-space neural renderer~\cite{gao2020deferred,philip2021free}, both methods learn to relight using buffers rendered with fixed, non-optimizable geometry. Due to the difficulty in performing an end-to-end, joint optimization of shape and appearance, the result quality of the above work is limited: once computed, errors in geometric estimation cannot be easily fixed, and may contaminate the subsequent appearance reconstruction.

On the other hand, few exception papers try to conduct a highly involved optimization that alternates between solving for shape and reflectance~\cite{wu2015simultaneous,Nam2018practical,10.1145/2980179.2980248}; the latter two even solve for unknown environment lighting. However, due to the non-differentiable nature of directly optimizing common traditional representations, approximations/tricks have to be applied from one place to another. Therefore, the result quality is still not satisfactory. It is not even clear if the optimization converges.

\subsection{Neural Relighting} 

With the advances in deep learning, neural implicit representations and/or modern large-scale optimization tools make it possible to jointly solve for geometry and appearance in a fully differentiable, end-to-end fashion. Compared with traditional relighting, direct optimization with respect to input photographs leads to higher quality results. Implicit representations like NeRF~\cite{mildenhall2020nerf} demonstrate unprecedented quality in novel view synthesis. And considerable research efforts are made to extend the idea to relighting~\cite{zhang2021nerfactor, Munkberg_2022_CVPR, sun2021nelf, bi2020neural}. Due to the space limit, below we briefly review representative approaches.

One class of existing work takes images under \emph{unknown} environment lighting(s) as input, and has to deal with the fundamental lighting-material ambiguity. These methods typically integrate approximate physical-based rendering (PBR), along with various regularization to better condition the optimization. Boss et al.~\shortcite{boss2021nerd} assume spatial coherence and jointly optimize a compressed latent BRDF space. Zhang et al.~\shortcite{zhang2021physg} employ a homogeneous specular appearance. Both methods ignore occlusion and indirect illumination. Extending from surface BRDFs, a MLP-predicted microflake volume is proposed in~\cite{zhang2023nemf}. Jin et al.~\shortcite{jin2023tensoir} calculate visibility from the volume transmittance in a Siamese radiance field, and consider second-bounce illumination. A pre-trained neural renderer is proposed in~\cite{liang2023envidr}, as a neural approximation of the explicitly PBR rendering equation. Lyu et al.~\shortcite{lyu2022neural} incorporate PBR prior by bootstrapping light transport modeling with synthesized OLAT images as training data, and refine the result with captured photographs. The shape and appearance are not optimized in tandem.

Another class of work directly takes photographs captured with \emph{known/calibrated} lighting conditions as input. Srinivasan et al.~\shortcite{srinivasan2021nerv} train MLPs to predict fields of volumetric density, surface normal, material parameters, intersection, and visibility, which are jointly optimized via inverse rendering. Yu et al.~\shortcite{yu2023osf} employ a neural scattering function that approximates radiance transfer from a distant light, with OLAT input images. Recently, Zeng et al.~\shortcite{zeng2023nrhints} improve over~\cite{gao2020deferred} with a neural implicit radiance representation, and add shadow and highlight hints to help a network to model high frequency light transport effects. A hybrid point-volumetric representation is proposed in concurrent work~\cite{chung2024differentiable} for efficient inverse rendering. Due to hard visibility thresholding, transparent/furry objects are not supported.

While most state-of-the-art neural techniques can produce high-quality results, both the training and rendering costs are substantially more expensive than, e.g., GS-based methods. And it is non-trivial to directly apply the ideas here to GS, due to the considerable differences between the representations.

\subsection{Gaussian-Splatting-Based Relighting} 

Recently, 3D Gaussian Splatting~\cite{kerbl3Dgaussians} introduces a highly efficient differentiable rasterization pipeline for a Gaussian-based representation, substantially improving the training time and runtime performance. Low-order SH is employed in each vanilla Gaussian to represent Lambertian-dominant appearance variations under a fixed environment lighting. Several approaches replace SH with higher-frequency functions to improve the view-dependent synthesis quality, which, however, cannot support lighting change \cite{yang2024specgaussian,malarz2023gaussian,ye2024gsdr}.

Towards the goal of relighting with more complex, general appearance, a number of techniques have been proposed. Similar to neural relighting, the majority of related work here takes images under an unknown environment lighting as input~\cite{jiang2023gaussianshader,shi2023gir,liang2023gsir,gao2023relightable}. The basic idea is to model the appearance for each 3D Gaussian as an isotropic parametric BRDF, precompute or ray-trace the visibility, sample indirect illumination and store as low-frequency SH, and perform inverse rendering. All these methods require well defined \emph{surface} normals to properly regularize their optimizations, which limits the applicable geometry to opaque ones with clear boundaries.

Another line of work takes images captured with varying light sources as input. Saito et al.~\shortcite{saito2023relightable} propose a relightable head avatar. For each 3D Gaussian, the specular reflectance is modeled as a single learnable isotropic spherical Gaussian, and the light visibility is computed from a neural network.

All the above work for general objects/scenes does not handle challenging appearance such as anisotropic reflections, with the exception of specialized models (e.g., ~\cite{luo2024gaussianhair} for hair). Their relighting quality is limited, when compared with latest neural relighting approaches (e.g.,~\cite{zeng2023nrhints}). In comparison, we present the first general GS-based relightable representation for complex geometry and appearance. Unlike the aforementioned work, we do not rely on any regularizations/strong priors in the optimization (e.g., we do not require well defined surface normals), which can be fragile in handling complex cases. Our quality is comparable to or higher than state-of-the-art neural relighting, while our computation is substantially more efficient, by exploiting the differentiable rasterization pipeline of GS.

\section{Preliminaries}\label{sec:Preliminaries}

Our pipeline builds upon the highly efficient GS~\cite{kerbl3Dgaussians}. Similarly, we represent the geometry with anisotropic 3D Gaussians (or \emph{spatial Gaussians} in this paper), whose density at a 3D location $\mathbf{p}$ is defined as:
\begin{equation}
    G_\mathrm{spa}(\mathbf{p})=\exp(-\frac{1}{2}(\mathbf{p}-\boldsymbol{\mu})^{\top} \Sigma^{-1}(\mathbf{p}-\boldsymbol{\mu})).
\end{equation}
Here $\boldsymbol{\mu}$ is the 3D center of the Gaussian, and $\Sigma$ is a covariance matrix $\Sigma = RSS^TR^T$, where $S$ is a scaling matrix and $R$ is a rotation matrix.

Each spatial Gaussian is associated with an opacity $\gamma_j$ and a color $\textbf{c}_j$. To generate an image, the spatial Gaussians are projected to the screen as 2D Gaussian splats. Then, for each pixel, its intersecting Gaussian splats are sorted and alpha-blended as follows:
\begin{align}
\boldsymbol{\zeta} &=\sum_{j} \mathbf{c}_j \beta_j \gamma_j T_j, \\
T_j &= \prod_{k=0}^{j-1} ( 1 - \beta_k \gamma_k ). 
\label{eqn:splat}
\end{align}
Here $\boldsymbol{\zeta}$ is the final color of the current pixel, $T_j$ is the cumulative opacity for the top-most $j$ Gaussian splats, and $\beta_k$ is the density at the current pixel center of the k-th Gaussian splat.

\section{Our Approach}
\label{sec:approach}

We take photographs of an object/scene from different calibrated views, lit with one point light at a time as input, and output a set of spatial Gaussians to represent the geometry, each of which is associated with an opacity and an appearance function, mainly represented as a linear combination of angular Gaussians.

\begin{figure*}[!htb]
    \includegraphics[width=\textwidth]{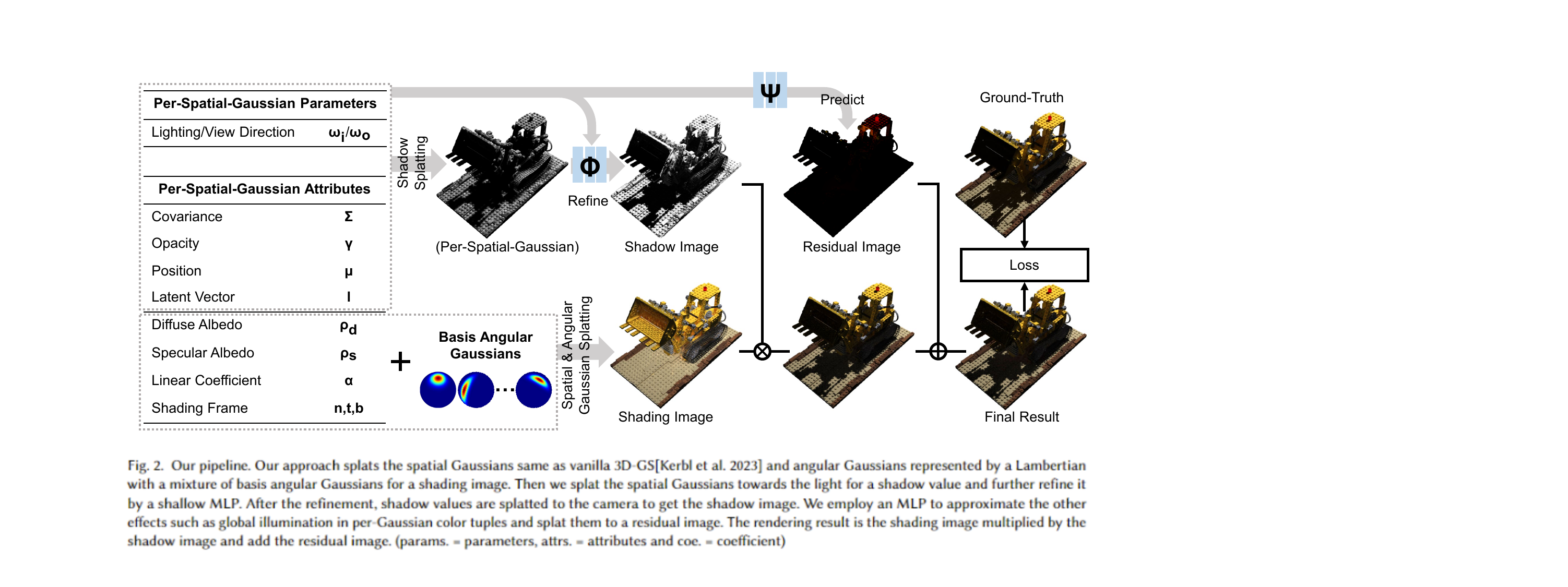}
    \caption{Our deferred-shading-based pipeline. First, we color each spatial Gaussian by evaluating its appearance function, defined as a Lambertian plus a linear combination of basis angular Gaussians, and splat into a shading image. Next, for each spatial Gaussian, we compute a shadow value by splatting all of them towards the light (i.e., shadow splatting), and refine it with an MLP. We color each spatial Gaussian with its shadow value, and splat them into a shadow image. Finally, we color each spatial Gaussian with another MLP that represents unhandled effects like global illumination, and splat them into a residual image. The final rendering result is computed by multiplying the shading image with the shadow one, and adding the residual image, on a per-pixel basis.
    }
    \label{fig:pipeline}
\end{figure*}

To efficiently render an image under a point light, a deferred shading approach is adopted. First, we color each spatial Gaussian by evaluating its appearance function, and splat them into a \emph{shading image} (\sec{sec:appearance}). Next, for each spatial Gaussian, we compute a shadow value by splatting all of them towards the light (which we call \emph{shadow splatting}), and refine it with an MLP. We color each spatial Gaussian with its own shadow value, and splat them into a \emph{shadow image} (\sec{sec:shadow}). Finally, we color  each spatial Gaussian with another MLP that represents unhandled effects like global illumination, and splat them into a \emph{residual image}~(\sec{sec:other effects}). The final rendering result is computed by multiplying the shading image with the shadow one, and adding the residual image, on a per-pixel basis. Please refer to~\figref{fig:pipeline} for a graphical illustration.

For training, an end-to-end, joint optimization of spatial Gaussians and corresponding appearance functions are performed. Similar to vanilla GS, it minimizes the differences between the input photographs and our rendering. Adaptive density control in GS is also applied.

\subsection{Shading} 
\label{sec:appearance}

To accurately represent complex appearance (e.g., anisotropic reflections), we replace for each spatial Gaussian the low-order SH as defined in vanilla GS with a view- and lighting-dependent function $f$ as:
\begin{equation}
f(\boldsymbol{\omega}'_i, \boldsymbol{\omega}'_o) = \rho_d f_d(\boldsymbol{\omega}_i') + \rho_s f_s(\boldsymbol{\omega}'_i, \boldsymbol{\omega}'_o).
\label{eqn:reflectance func}
\end{equation}
Here $\rho_d$/$\rho_s$ is the diffuse/specular albedo, $f_d$/$f_s$ is the diffuse/specular appearance function to be defined in the remainder of this subsection, and $\boldsymbol{\omega}'_i$/$\boldsymbol{\omega}'_o$ is the local lighting/view direction, respectively. The learnable shading frame at a spatial Gaussian is defined as [$\mathbf{n}$, $\mathbf{t}$, $\mathbf{b}$], which, as the name suggests, consists of normal, tangent and binormal. In practice, we do not explicit store the three vectors. Instead, we use a unit quaternion to express an equivalent rotation transform from the world space to the shading frame of a particular Gaussian. Note that the shading frame is totally independent from the 3 axes of the spatial Gaussian (represented in $\Sigma$).

To generate a shading image, we color each spatial Gaussian by evaluating $f$, and then perform splatting. Below we describe the details about $f_d$ and $f_s$.

\paragraph{Diffuse Appearance.} 
We modify a common cosine-weighted Lambertian function to the following function $f_d$:
\begin{equation}
    f_d(\boldsymbol{\omega}'_i) = \frac{\mathrm{ELU}(\mathbf{n}' \cdot \boldsymbol{\omega}'_i) + \varepsilon(1-\frac{1}{e})}{(1 + \varepsilon(1-\frac{1}{e})) \pi}.
    \label{eqn:diffuse}
\end{equation}
Here $\mathbf{n}'$ is the normal in the shading frame, and we set the hyperparameter of ELU and $\varepsilon$ as 0.01. Note that $f_d$ is slightly different from the standard definition, and  its gradient is non-zero for any $\boldsymbol{\omega}'_i$, which is amenable for differentiable optimization. In comparison, the original cosine-weighted Lambertian has a zero gradient over the lower hemisphere, which would form a "dead zone" once the optimization gets stuck there.

\paragraph{Specular Appearance.} To represent complex all-frequency specular appearance, we model $f_s$  as a mixture of modified anisotropic spherical Gaussians (denoted as \emph{angular Gaussians} in this paper) below:
\begin{equation}
    f_s(\boldsymbol{\omega}'_i, \boldsymbol{\omega}'_o) = \sum_{j}{\alpha_j G_{\mathrm{ang},j} (\mathbf{h'}) },
    \label{eqn:hight_frequency_func}
\end{equation}
where $\alpha_j$ is a weight, and $\mathbf{h'}$ is the half vector computed as $\mathbf{h'} = \frac{\boldsymbol{\omega}'_i + \boldsymbol{\omega}'_o}{\lVert\boldsymbol{\omega}'_i + \boldsymbol{\omega}'_o \rVert}$. $G_{\mathrm{ang},j}$ is an angular Gaussian, defined as:
\begin{equation}
    G_{\mathrm{ang}}(\mathbf{h}') = \frac{1}{\sigma_z} \exp \left(-\frac{1}{2}{\left(\frac{\arccos(\mathbf{h}' \cdot \mathbf{z}) \sqrt{(\frac{\mathbf{s}' \cdot \mathbf{x}}{{\sigma}_x})^2 + (\frac{\mathbf{s}' \cdot \mathbf{y}}{{\sigma}_y})^2}}{{\sigma}_z}\right)}^2 \right).
    \label{eqn:our_asg}
\end{equation}
Here [$\mathbf{x}$, $\mathbf{y}$, $\mathbf{z}$] is the local frame of the angular Gaussian,
$\mathbf{s}'$ is the normalized result of the projection of $\mathbf{h}'$ onto the $\mathbf{x}$-$\mathbf{y}$ plane, and ${\sigma}_x$/${\sigma}_y$/${\sigma}_z$ are the standard deviations in three dimensions. Note that we extend the isotropic definition in~\cite{saito2023relightable} with~\cite{2013asg} to support anisotropy. Directly employing the definition from~\cite{2013asg} often cannot model highly specular reflections well, and its smooth term is unfriendly to differentiable optimization, according to~\cite{saito2023relightable} and our pilot study. We call the evaluation of~\eqnref{eqn:hight_frequency_func} as \emph{angular Gaussian splatting}, as it involves the mixing of multiple Gaussians.

Furthermore, for a particular object/scene, a set of the basis angular Gaussians are shared across all spatial Gaussians, essentially exploiting the spatial coherence to better condition the optimization, as common in related work~\cite{lensch2003image, chen2014reflectance, Nam2018practical}. Consequently, for each spatial Gaussian, the complete set of learnable parameters to represent an $f$ consist of [$\mathbf{n}$, $\mathbf{t}$, $\mathbf{b}$], [$\mathbf{x}$, $\mathbf{y}$, $\mathbf{z}$], [${\sigma}_x$, ${\sigma}_y$, ${\sigma}_z$], $\rho_d$, $\rho_s$, and the set of weights \{$\alpha_j$\} to linearly combine the shared basis angular Gaussians, whose total number is 8 in main experiments. Note that when the input appearance information is sufficient to condition our optimization~\cite{tunwattanapong2013acquiring, ma2021free}, it is possible to use a separate set of angular Gaussians for each spatial Gaussian, instead of sharing them, to further improve the result quality.

\subsection{Shadowing}
\label{sec:shadow}

\begin{figure}[!htb]
    \begin{minipage}{\linewidth}
        \centering
        \begin{minipage}[t]{\linewidth}
        \centering
            \includegraphics[width=\linewidth]{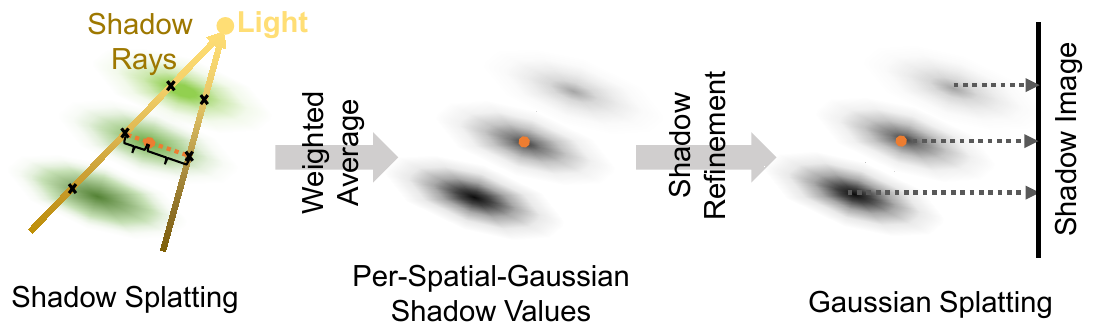}
            \put(-233,35){\fontsize{4.5}{5}\selectfont \textbf{$\beta_{m_{1}}=0.6$}}
            \put(-212,31){\fontsize{4.5}{5}\selectfont \textbf{$\beta_{m_{2}}=0.3$}}
            \put(-228,46){\fontsize{4.5}{5}\selectfont \textbf{$T_{m_{1}}=0.4$}}
            \put(-206,42){\fontsize{4.5}{5}\selectfont \textbf{$T_{m_{2}}=0.6$}}
            \put(-147,47){\fontsize{4.5}{5}\selectfont \textbf{$T=\frac{\sum_{m}T_m\beta_m}{\sum_{m}\beta_m}=0.42$}}
            \put(-54,47){\fontsize{4.5}{5}\selectfont \textbf{$T'=0.40$}}
        \end{minipage}
    \end{minipage}
    \caption{Shadow computation. We first splat all spatial Gaussians towards the light source, and compute the accumulated opacity $T_m$ and the density ${\beta}_m$ at each intersection with each shadow ray (left). For each spatial Gaussian, all of its ${T}_m$ with respect to different shadow rays are weighted-averaged by corresponding ${\beta}_m$ to obtain a shadow value $T$ (center). Next, this shadow value is refined with a small MLP. Finally, Gaussians with refined shadow values ($T'$) are splatted towards the camera to produce the shadow image (right). The above Gaussians may appear incorrectly, when viewed in a program other than Adobe Acrobat.}
    \label{fig:shadow}
\end{figure}
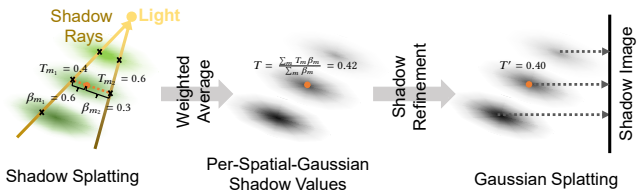

The shading image does not account for any shadowing effects. To support them, a na\"ive but expensive method would trace a shadow ray from a scene point to the point light, and perform a line integral of opacity along the way as the final visibility result with respect to the light. To develop an efficient algorithm for shadow computation on Gaussians, we observe that traditional shadow mapping~\cite{williams1978casting} reuses the high-performance rasterizaiton pipeline for the view from the light instead of the camera. Here we apply a similar idea, by performing high-performance Gaussian splatting towards the point light. Please see~\figref{fig:shadow} for an illustration.

Specifically, we first splat all spatial Gaussians towards the point light, by setting up a perspective camera whose center is the light, similar to how a shadow map is computed (i.e., shadow splatting). Here we use the same image resolution as an input image. For a shadow ray (with an index of $m$), a number of spatial Gaussians whose 2D splats intersecting the ray are sorted, according to the distances to the light. Next, we compute the cumulative opacity $T_m$, according to~\eqnref{eqn:splat}, for an intersecting spatial Gaussian as its shadow value.

Note that the 2D splat of a spatial Gaussian is likely to intersect a number of shadow rays, with multiple shadow values computed at each ray. In this case, we compute an average $T = \frac{\sum_{m}{\beta_{m} T_{m}}}{\sum_{m}{\beta_{m}}}$ as the result, weighted by $\beta_m$, the projected 2D Gaussian density at an intersection. 
Similar to shadow mapping, we apply a shadow bias of 0.015 to alleviate "z-fighting".

To further improve the shadow quality (e.g., denoise), we refine for each spatial Gaussian its shadow value with an MLP, $\Phi$, as follows:
\begin{equation}
T' = \Phi(T, \boldsymbol{\omega}_i; \boldsymbol{\mu}, \mathbf{l}).
\end{equation}
Here $T/T'$ is the shadow value before/after neural refinement, respectively. In addition, $\Phi$ is a small 3-layer MLP with \{32, 32, 32\}. Each hidden layers is followed by a leaky ReLU activation, while the output layer a sigmoid one. The parameter $\boldsymbol{\mu}$ (i.e., the spatial Gaussian center) is designed to make the MLP spatial aware, and the learnable $\mathbf{l}$ is a per-spatial-Gaussian 6D latent vector. We apply 4-band positional encoding to $\boldsymbol{\mu}$ and $\boldsymbol{\omega}_i$ before sending to $\Phi$. Finally, we color each spatial Gaussian with its refined shadow value, and splat into a shadow image.

\subsection{Other Effects}
\label{sec:other effects}

To consider other light transport not modeled in \sec{sec:appearance}~\&~\ref{sec:shadow} (e.g., global illumination), we employ another MLP, $\Psi(\boldsymbol{\omega}_o; \boldsymbol{\mu}, \mathbf{l})$, to predict the impact of these effects for each spatial Gaussian. Specifcically, $\Psi$ is a 3-layer MLP with \{128, 128, 128\}. Each hidden layer is followed by a leaky ReLU activation, while the output layer a sigmoid one. The MLP is a function of $\boldsymbol{\omega}_o$ only, which is a common parameterization for representing indirect illumination in real-time rendering~\cite{10.5555/3294495}. In addition, the parameter $\boldsymbol{\mu}$ is the spatial Gaussian center, and the learnable $\mathbf{l}$ is a shared latent vector defined in~\sec{sec:shadow}. We apply 4-band positional encoding to $\boldsymbol{\mu}$ and $\boldsymbol{\omega}_o$. Finally, we evaluate $\Psi$ to color each spatial Gaussian, and splat into a residual image.

\subsection{Training}\label{sec:training}

\paragraph{Loss.} We use the loss function from~\cite{kerbl3Dgaussians}, defined as follows:
\begin{equation}
    \mathcal L = (1 - \lambda) \mathcal L_1 + \lambda \mathcal L_{\mathrm{D-SSIM}}.
    \label{eqn:loss}
\end{equation}
Here $\mathcal L_1$ is the L1 image loss, $\mathcal L_\mathrm{D-SSIM}$ is the sum of SSIM~\cite{wang2004image} of each difference image between an input photograph and our corresponding rendering, and $\lambda$ = 0.2 in all experiments. Note that we do not impose any regularization on intermediate results/components. We find it sufficient and elegant to use the above end-to-end image loss for high-quality relighting.

\paragraph{Initialization.} We initialize geometry properties of spatial Gaussians, following vanilla GS. For an angular Gaussian, we randomly sample $\sigma_z$ from [0.13, 0.69], and set $\sigma_x=0.5$ and $\sigma_y=1.0$. Both $\rho_d$ and $\rho_s$ are initialized as $(1,1,1)$, and each $\alpha$ is set to 0.5. The local/shading frame of each spatial/angular Gaussian is initialized to align with the axes in the world space.

\paragraph{Training Strategy \& Details.} To reduce the probability of getting stuck with an undesired local minimum and to increase robustness in practice, we use a two-stage training strategy to gradually increase the degrees of freedom. First, we limit the appearance function to be the Lambertian term only, and train for 15K iterations. We find that the shading frames converge stably in this stage. Next, we use the full appearance function with specular reflections, and train for 100K iterations. In all experiments, we employ the Adam optimizer with a momentum of 0.9. The learning rate varies with different parameters, similar to~\cite{kerbl3Dgaussians}. For $\rho_d$ and $\rho_s$, the learning rate is set to 0.01. For angular Gaussians, we use a fixed rate of 0.01 before 40K iterations, and exponetially decay it to 0.0001 at 90K, and fix it afterwards.

\subsection{Rendering}\label{sec:rendering}

For a given view and a point light, the rendering process with our representation is described in the beginning of~\sec{sec:approach}. To support a directional light, we switch from perspective projection to orthographic one in the shadow splatting process. Furthermore, rendering with an environment light is implemented as a linear combination of the results under a number of sampled directional lights.

\section{Results \& Discussions}

All experiments are conducted on a workstation with dual AMD EPYC 7763 CPUs, 768GB DDR4 memory and an NVIDIA GeForce RTX 4090 GPU.  It takes 40-70 minutes to train our representation (120K-750K spatial Gaussians and 8 basis angular Gaussians), whose rendering speed is over 90fps.

We reconstruct objects/scenes with a wide variation in geometry (from solid to fluffy) and appearance (from translucent to anisotropic) from multi-view point-lit images. Four forms of input data are tested: (1) rendered images of syntheic NeRF~\cite{mildenhall2020nerf}; (2) rendered images of captured results from~\cite{kang2019learning} and OpenSVBRDF~\cite{ma2023open}; 
(3) captured point-lit photographs from NRHints~\cite{zeng2023nrhints}; and (4) multi-view photometric images captured by a professional lightstage. The input images for (1) and (2) have a spatial resolution of $512^2$, while the resolution is $512^2$ or $1024^2$ for (3) and (4).

For comparison experiments, we take the official code of each method and retrain with the same set of 500-2,000 input images. For quantitative assessments of reconstruction quality, we compute PSNR, SSIM, and LPIPS~\cite{zhang2018unreasonable} averaged over all test images. Please also refer to the accompanying video for animated results with varying view and lighting. For ablation experiments, please refer to the supplemental material, as the space of the main paper is limited.

\subsection{Results}

\paragraph{Synthetic.} \figref{fig:synthetic-evaluation} and \figref{fig:pointlight_comparison} show our reconstruction results from rendered images of synthetic/captured subjects. For each subject, we use 2,000 training images and 400 test ones, with randomly sampled view and point light. {\sc{Translucent}} exhibits complex subsurface scattering; {\sc{AnisoMetal}} and {\sc{Drums}} contain strong anisotropic appearance; {\sc{FurBall}} has a fuzzy geometry; and {\sc{Lego}} contains complex occlusions and shadows. In addition, {\sc{MaterialBalls}}, {\sc{Tower}}, {\sc{Fabric}}, {\sc{Cup}} and {\sc{Egg}} show complex, spatially-varying specular reflections. In all cases, we successfully reconstruct a wide variety of challenging shapes and appearance with our representation.

\paragraph{Captured.} In \figref{fig:real evaluation} and \figref{fig:pointlight_comparison}, we generate realistic novel lighting-and-view synthesis results from the data of~\cite{zeng2023nrhints}, acquired with a handheld camera and a smartphone with a flash. Exactly the same training and test data from their paper are used. {\sc{Cup-Fabric}} consists of translucent materials (cup), and isotropic (balls) and anisotropic reflections (fabric); {\sc{Pixiu}} shows strong subsurface scattering and self-occlusions; {\sc{Fish}} ,{\sc{Cluttered}} and {\sc{Cat}} contain intricate details, like shadow and glint (ground-plane) and complex appearance (fur). {\sc{Pikachu}} includes glossy highlights and considerable self-occlusions.

In Fig.~\ref{fig:real_ours}, we reconstruct from multi-view photometric photographs, captured with a professional lightstage with 24,576 LEDs and 2 cameras, similar to~\cite{kang2019learning}. For each subject, we use 2,000 training images and 400 test images. {\sc{Zhaojun}} and {\sc{Boot}} contain furry geometry and complex glinty/anisotropic appearance; {\sc{Fox}} has highly complex occlusions and reflections on grass-like ground and the hair; {\sc{Li‘lOnes}} is a doll with challenging parts like long hair, which often requires special, separate handling in existing work~\cite{saito2023relightable}. {\sc{Container}} and {\sc{Nefertiti}} include highly specular reflections with complex spatial variations. For all these highly challenging cases, we demonstrate high-quality reconstructions with a unified representation.

\begin{figure}[htb]
    \includegraphics[width=\linewidth]{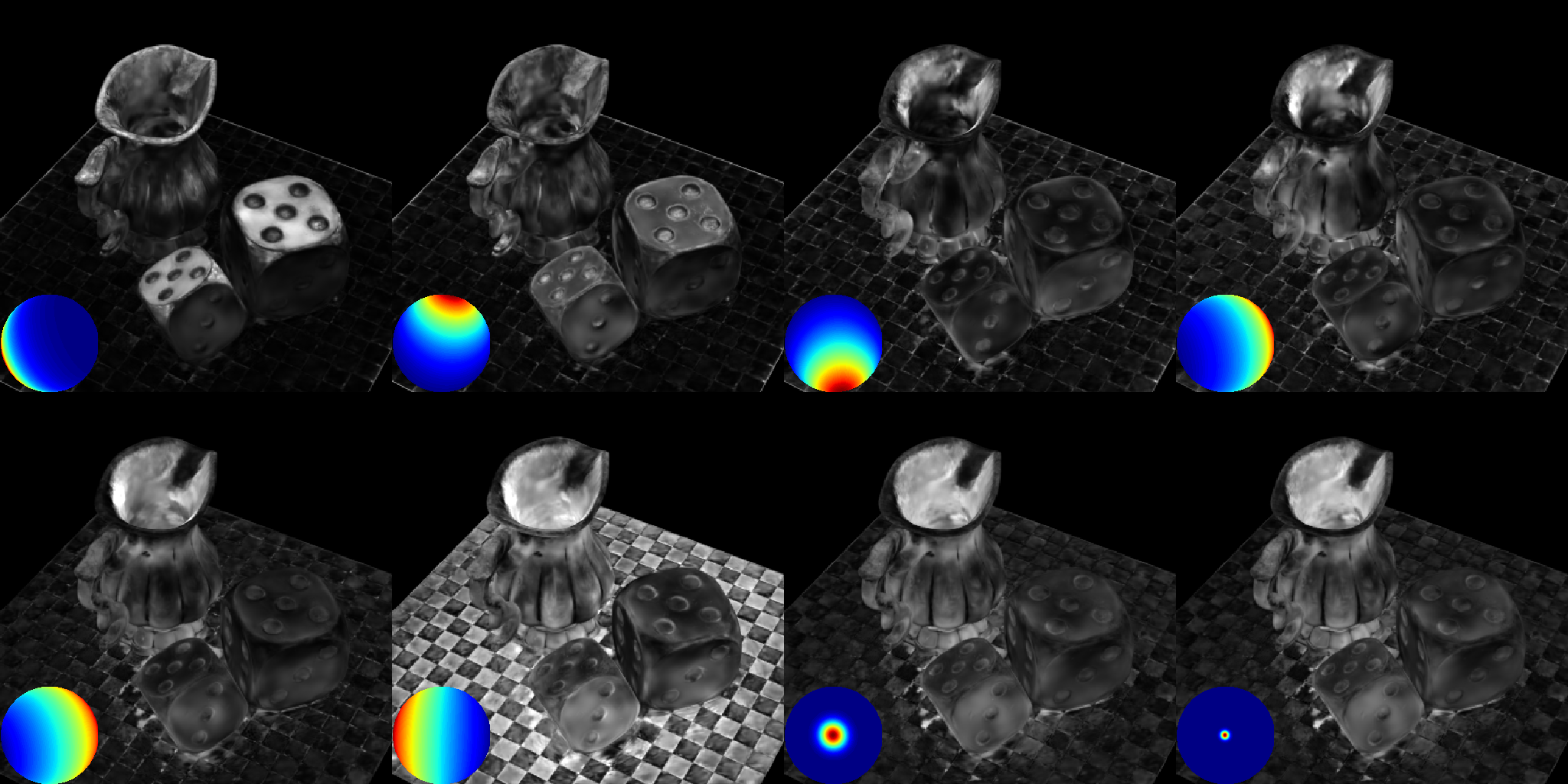}
    \caption{Visualization of the basis angular Gaussians and the spatial distributions of corresponding weights. Each image shows the spatial distribution of the weight for a particular basis angular Gaussian, which is color-coded in the bottom-left corner. A brighter pixel indicates a larger weight.}
    \label{fig:basisASG}
\end{figure}

\paragraph{Visualizations.} In \figref{fig:decompose_vis}, we visualize the components of the pipeline, for a better understanding of what is going on under the hood. Note that the intermediate results demonstrate decent quality/decoupling from each other, despite that we do \emph{not} impose regularization/constraint on any of them, unlike the majority of GS-based work. This shows the elegance of our simple, end-to-end image loss. In addition, \figref{fig:basisASG} visualizes the basis angular Gaussians and the spatial distributions of corresponding weights (via splatting) of the same scene.

\input{figure/figure_only/real_ours}
\input{figure/figure_only/eval_nrhints}

\subsection{Comparisons} \label{sec:comparsion}

In Fig.~\ref{fig:NRTF_comparison}, we compare with Neural Radiance Transfer Fields~\cite{lyu2022neural}. Both approach are trained with the same set of 2,000 directional-lit images. Our results show higher-quality shadows and specular reflections than theirs. Moreover, due to the surface-based appearance representation, they cannot work well on fuzzy geometry like {\sc{FurBall}}.

Fig.~\ref{fig:pointlight_comparison} compares with NRHints~\cite{zeng2023nrhints}, the state-of-the-art relightable implicit representation. We can reconstruct the anisotropic highlights in {\sc{Drums}} and the specular reflections on the floor of {\sc{Fish}}, which their approach struggles with. For {\sc{Fur}}, {\sc{Lego}} and {\sc{Cat}}, they fail to reproduce many of the original spatial details, although they sometimes achieve a higher score than ours with the blurry reconstructions. Compared with NRHints, we obtain higher-quality or comparable results, and more than \emph{an order of magnitude} higher performance both in training (40-70min vs. 15 hrs) and rendering  (90fps vs. <1fps) on the same workstation.

In Fig.~\ref{fig:comparsion_osf}, we compare with OSF~\cite{yu2023osf}, one state-of-the-art method for reconstructing sub-surface scattering appearance, on {\sc{Translucent}}. While not explicitly modeled, scattering effects are faithfully recovered with our representation. In comparison, the lack of self-occlusion handling and accurate surface reflectance in OSF leads to a lower-quality result.
\input{figure/figure_only/cmp_OSF}

In Fig.~\ref{fig:envlight_comparison}, we compare with GS based/NeRF-like relighting methods, which take images lit with unknown environment lighting as input, including GaussianShader~\cite{jiang2023gaussianshader}, GS-IR~\cite{liang2023gsir}, Relightable 3D Gaussian~\cite{gao2023relightable} and Tenso-IR~\cite{jin2023tensoir}. For these methods, we train with the same set of 2,000 images rendered with an environment map, and test under a different one. To be as fair as possible, the same ground-truth environment map is supplied to all alternative methods as input during training; our approach uses the same number of point-lit input images; Another test environment map is used for relighting. Our quality clearly surpasses alternative approaches in all cases. This is not surprising: state-of-the-art relighting techniques (e.g.,~\cite{zeng2023nrhints}) take images with \emph{known} lighting as input, rather than with unknown environment lighting, to produce high-quality results.

\section{Limitations \& Future Work}

Our work is subject to a number of limitations. First, we do not consider transparent materials, such as glass or gems. It will be interesting to replace the MLP $\Psi$ with an explicitly modeling of other light transports like refraction or internal reflection. Moreover, under certain lighting/view conditions, our shadows are not as crisp as, e.g., using a mesh-based representation for geometry. Also for extremely high-frequency anisotropic appearance, the reconstructed highlights might blink. The main cause for both cases is the insufficient granularity of spatial Gaussians. We are intrigued to develop more advanced density control method, as well as establish additional direct gradient pathway(s), to solve this problem.

In the future, it is promising to search for the optimal acquisition conditions (i.e., view/lighting) for our representation to reduce the number of input images and improve the reconstruction quality at the same time, by exploiting, e.g., the highly efficient learned illumination multiplexing~\cite{kang2021neural}. While the images we capture with a professional lightstage is already useful as a benchmark for future research, it is desirable to build a large-scale database in the spirit of~\cite{ma2023open} to facilitate generative tasks like ~\cite{zeng2024dilightnet,Poirier2024diffusionrelighting} based on our representation.

\begin{acks}
We would like to express our gratitude to Yue Dong, Guojun Chen, Xianmin Xu and Yaowen Chen for their generous help to this project. This work is partially supported by NSF China (62332015, 62227806 \& 62421003), the Fundamental Research Funds for the Central Universities (226-2023-00145), the XPLORER PRIZE, and Information Technology Center and State Key Lab of CAD\&CG, Zhejiang University.
\end{acks}

\clearpage
\bibliographystyle{ACM-Reference-Format}
\bibliography{RGS}

\appendix
\input{src/RGS-figure-only}

%% file: figure/figure_only/real_ours.tex
\begin{figure*}[htb]
    \begin{minipage}{\textwidth}
        \centering
        \begin{minipage}{1.15in}
            \centering
            \scalebox{.85}{Ground-Truth}
        \end{minipage}
        \begin{minipage}{1.15in}
            \centering
            \scalebox{.85}{Ours}
        \end{minipage}
        \begin{minipage}{1.15in}
            \centering
            \scalebox{.85}{Ground-Truth}
        \end{minipage}
        \begin{minipage}{1.15in}
            \centering
            \scalebox{.85}{Ours}
        \end{minipage}
        \begin{minipage}{1.15in}
            \centering
            \scalebox{.85}{Ground-Truth}
        \end{minipage}
        \begin{minipage}{1.15in}
            \centering
            \scalebox{.85}{Ours}
        \end{minipage}
    \end{minipage}

    \begin{minipage}{\textwidth}
        \centering
        \begin{minipage}{1.15in}
            \includegraphics[width=\textwidth]{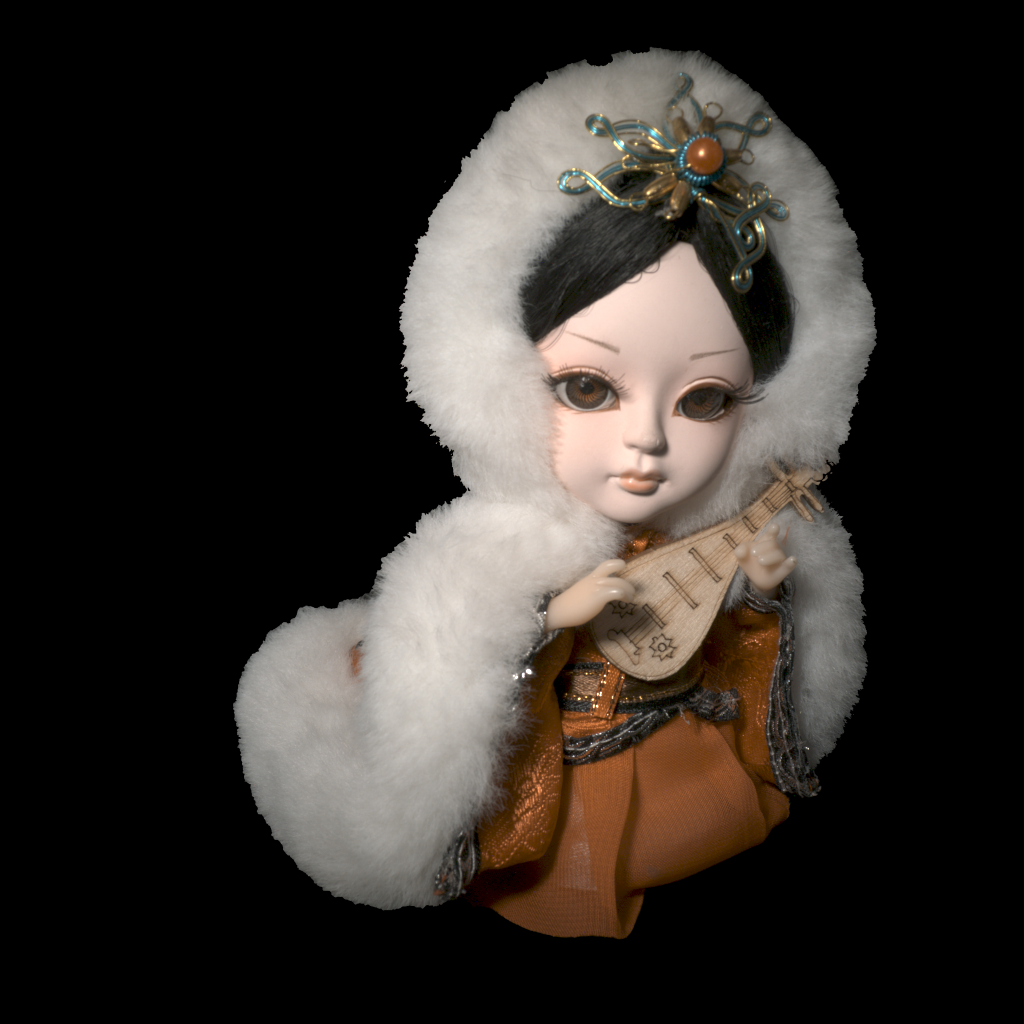}
        \end{minipage}
        \begin{minipage}{1.15in}
            \includegraphics[width=\textwidth]{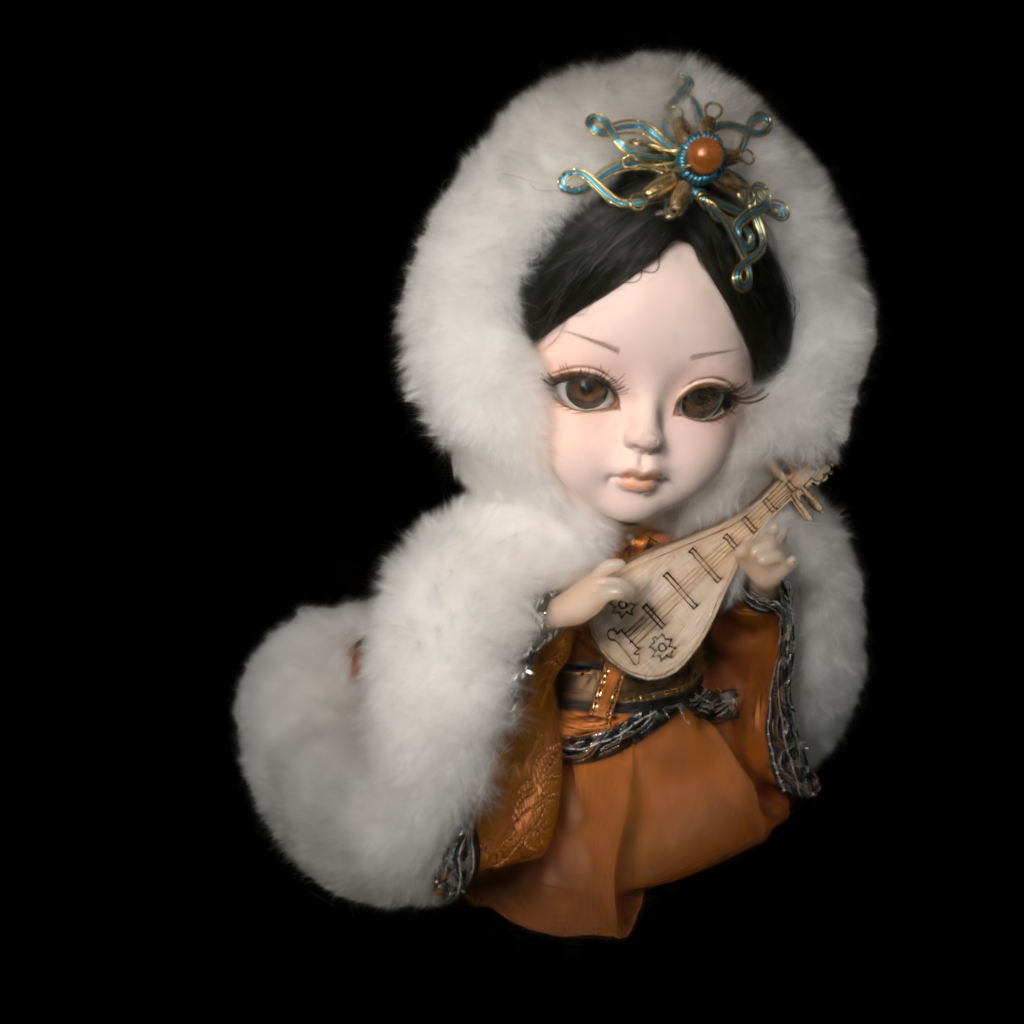}
        \end{minipage}
        \begin{minipage}{1.15in}
            \includegraphics[width=\textwidth]{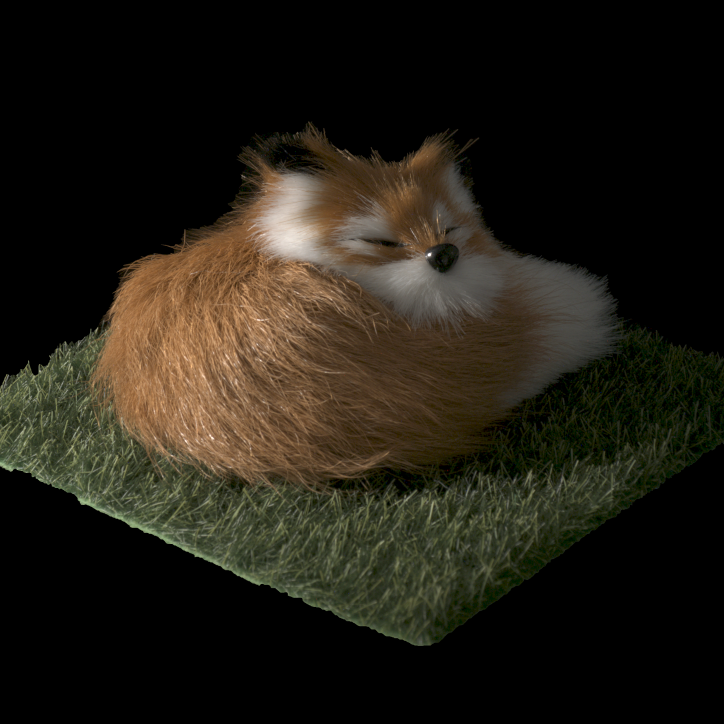}
        \end{minipage}
        \begin{minipage}{1.15in}
            \includegraphics[width=\textwidth]{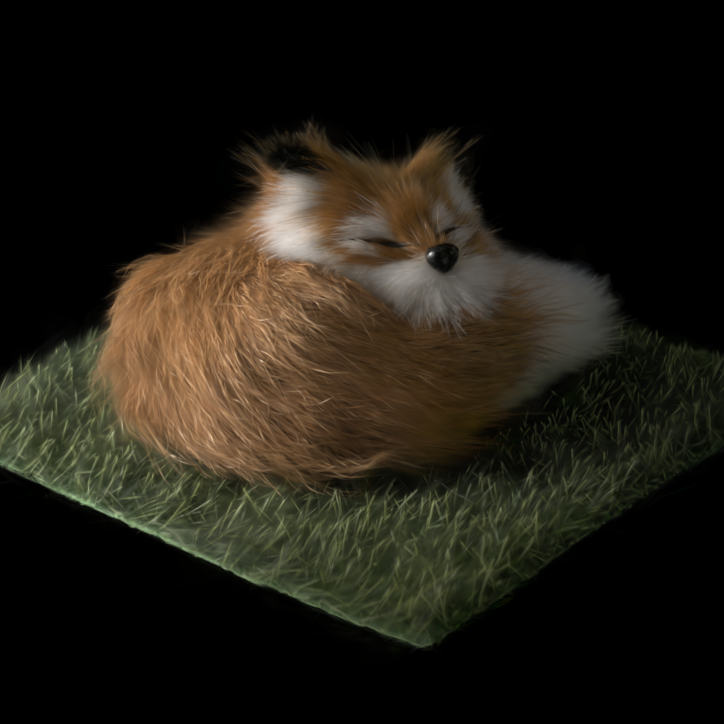}
        \end{minipage}
        \begin{minipage}{1.15in}
            \includegraphics[width=\textwidth]{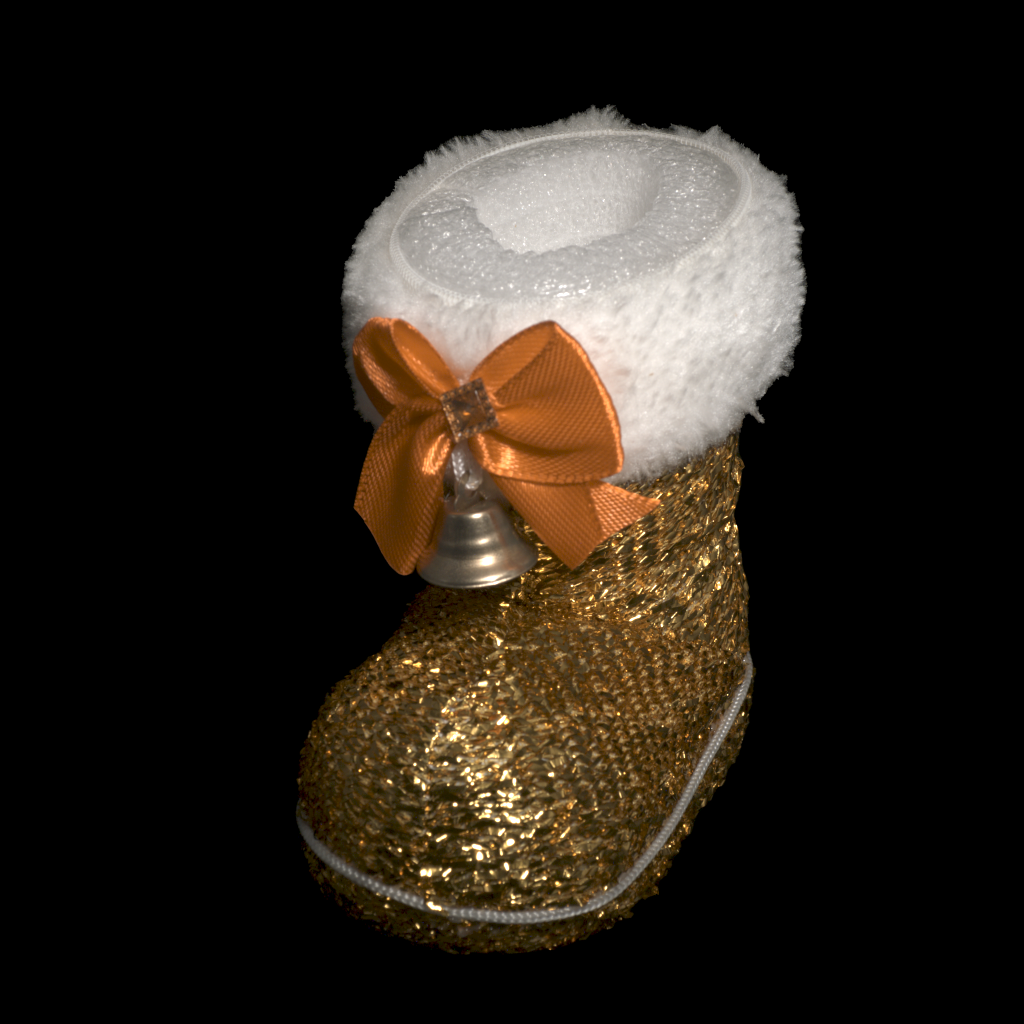}
        \end{minipage}
        \begin{minipage}{1.15in}
            \includegraphics[width=\textwidth]{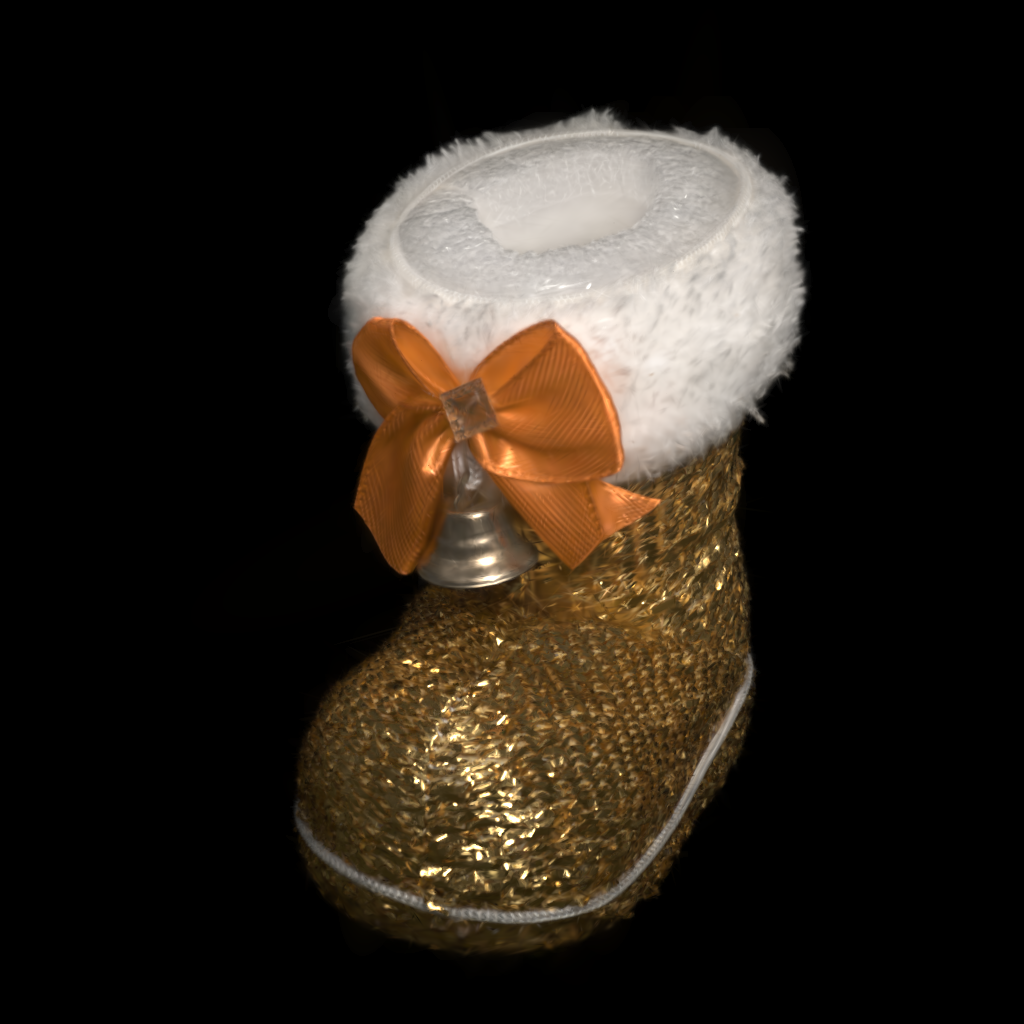}
        \end{minipage}
    \end{minipage}

    \begin{minipage}{\textwidth}
        \centering
        \begin{minipage}{2.2in}
            \centering
            \scalebox{.85}{\sc{Zhaojun: 0.9321 | 32.08 | 0.1071}}
        \end{minipage}
        \begin{minipage}{2.2in}
            \centering
            \scalebox{.85}{\sc{Fox: 0.9225 | 34.21 | 0.0745}}
        \end{minipage}
        \begin{minipage}{2.2in}
            \centering
            \scalebox{.85}{\sc{Boot: 0.898 | 28.84 | 0.1013}}
        \end{minipage}
    \end{minipage}

    \begin{minipage}{\textwidth}
        \centering
        \begin{minipage}{1.15in}
            \includegraphics[width=\textwidth]{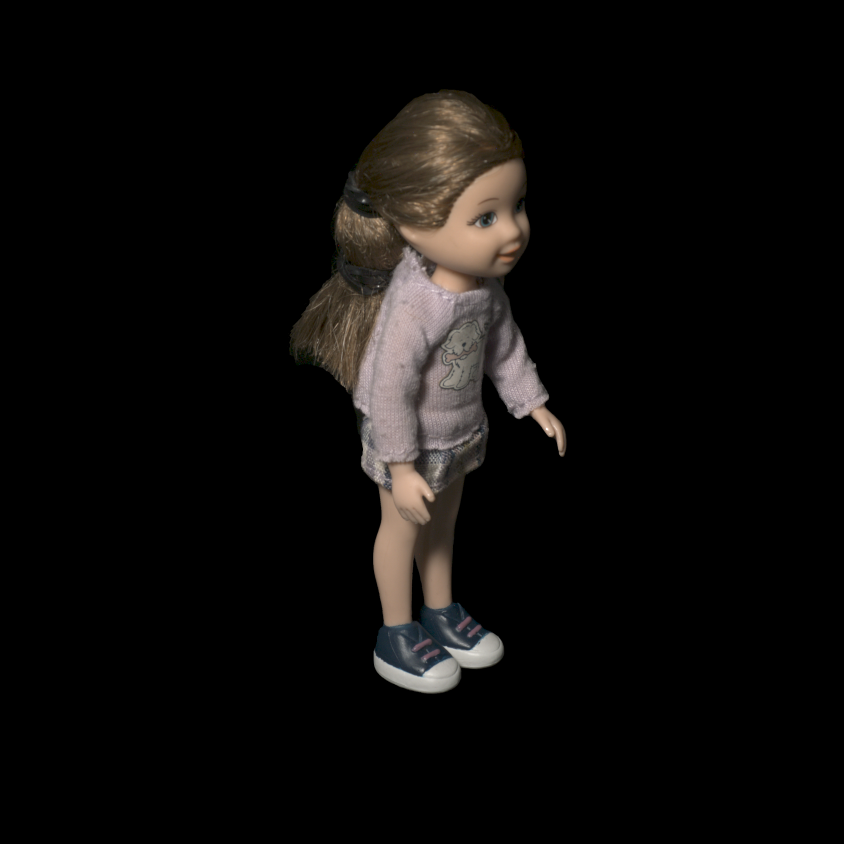}
        \end{minipage}
        \begin{minipage}{1.15in}
            \includegraphics[width=\textwidth]{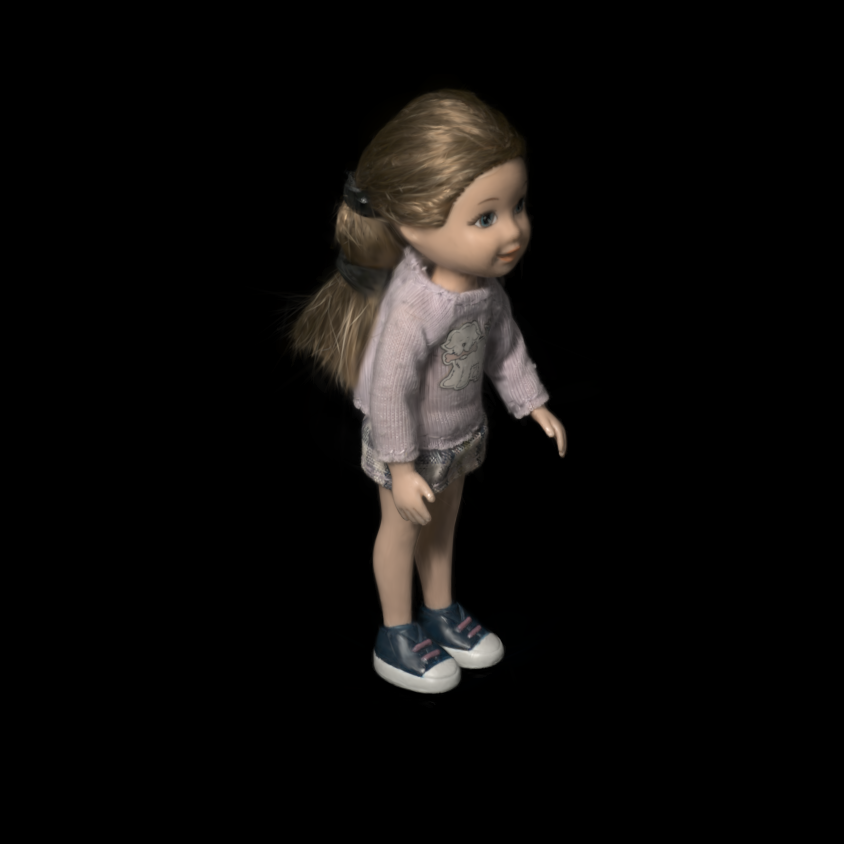}
        \end{minipage}
        \begin{minipage}{1.15in}
            \includegraphics[width=\textwidth]{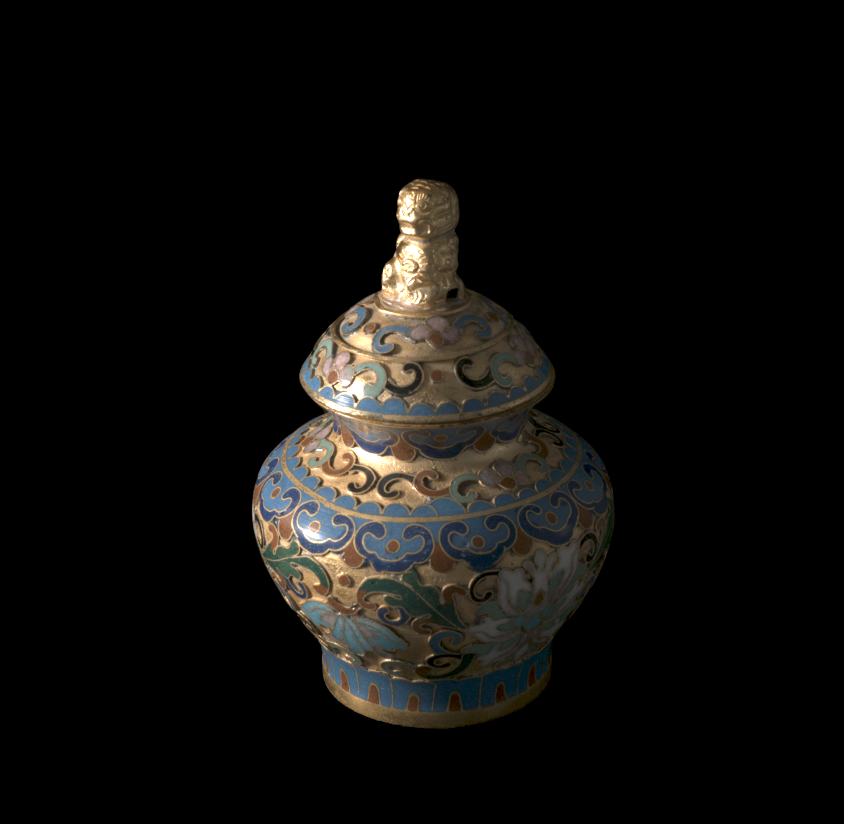}
        \end{minipage}
        \begin{minipage}{1.15in}
            \includegraphics[width=\textwidth]{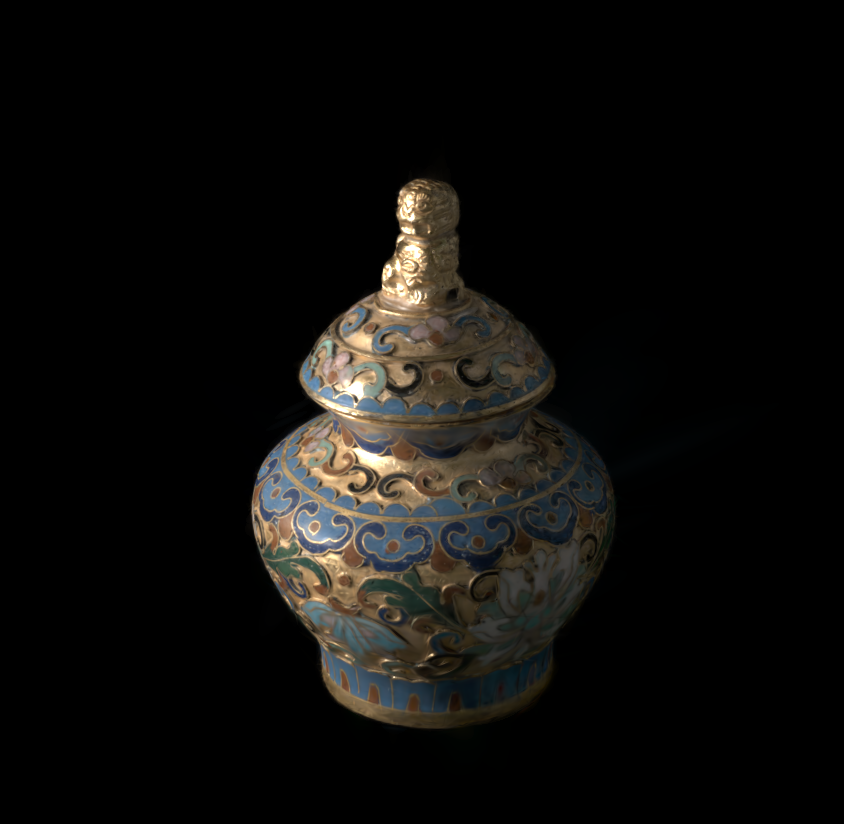}
        \end{minipage}
        \begin{minipage}{1.15in}
            \includegraphics[width=\textwidth]{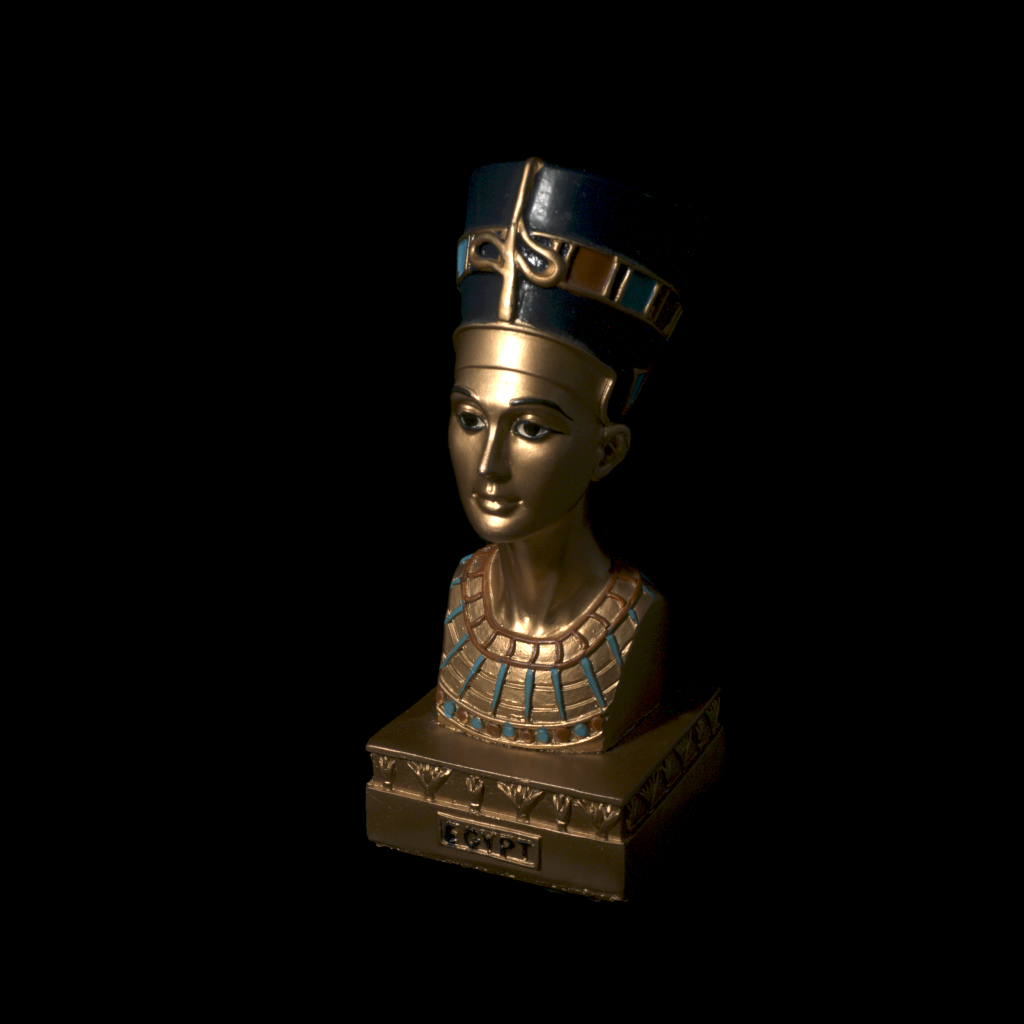}
        \end{minipage}
        \begin{minipage}{1.15in}
            \includegraphics[width=\textwidth]{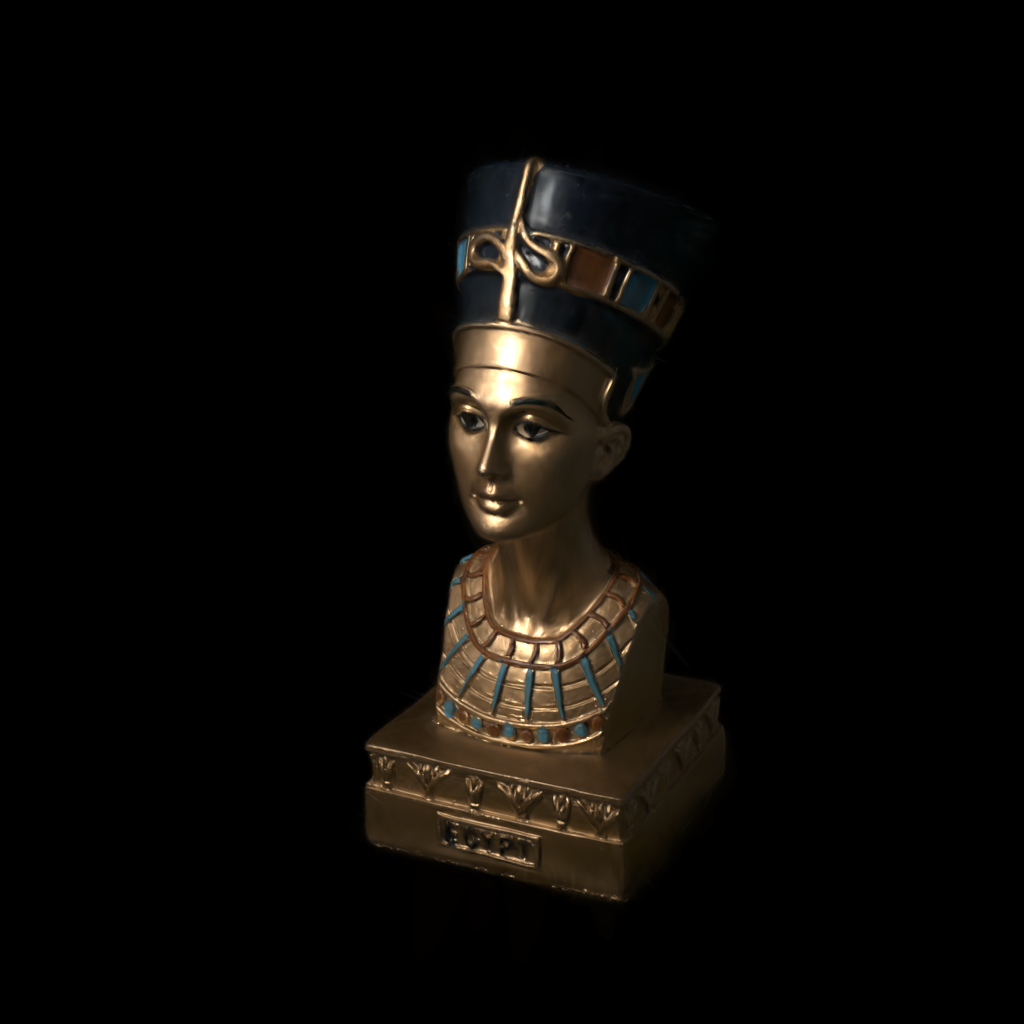}
        \end{minipage}
    \end{minipage}

    \begin{minipage}{\textwidth}
        \centering
        \begin{minipage}{2.2in}
            \centering
            \scalebox{.85}{\sc{Li'lOnes: 0.9809 | 38.61 | 0.0182}}
        \end{minipage}
        \begin{minipage}{2.2in}
            \centering
            \scalebox{.85}{\sc{Container: 0.9745 | 36.65 | 0.016}}
        \end{minipage}
        \begin{minipage}{2.2in}
            \centering
            \scalebox{.85}{\sc{Nefertiti: 0.956 | 36.58 | 0.0434}}
        \end{minipage}
    \end{minipage}
    
    \caption{Our relighting results on captured data from a professional lightstage. For each pair of images, the left one is the ground-truth photograph, and the right is our result. Average errors in SSIM, PSNR and LPIPS are reported at the bottom.}
    \label{fig:real_ours}
\end{figure*}

%% file: figure/figure_only/eval_nrhints.tex
\begin{figure*}[htb]
    \begin{minipage}{\textwidth}
        \centering
        \begin{minipage}{1.15in}
            \centering
            \scalebox{.85}{Ground-Truth}
        \end{minipage}
        \begin{minipage}{1.15in}
            \centering
            \scalebox{.85}{Ours}
        \end{minipage}
        \begin{minipage}{1.15in}
            \centering
            \scalebox{.85}{Ground-Truth}
        \end{minipage}
        \begin{minipage}{1.15in}
            \centering
            \scalebox{.85}{Ours}
        \end{minipage}
        \begin{minipage}{1.15in}
            \centering
            \scalebox{.85}{Ground-Truth}
        \end{minipage}
        \begin{minipage}{1.15in}
            \centering
            \scalebox{.85}{Ours}
        \end{minipage}
    \end{minipage}

    \begin{minipage}{\textwidth}
        \centering
        \begin{minipage}{1.15in}
            \includegraphics[width=\textwidth]{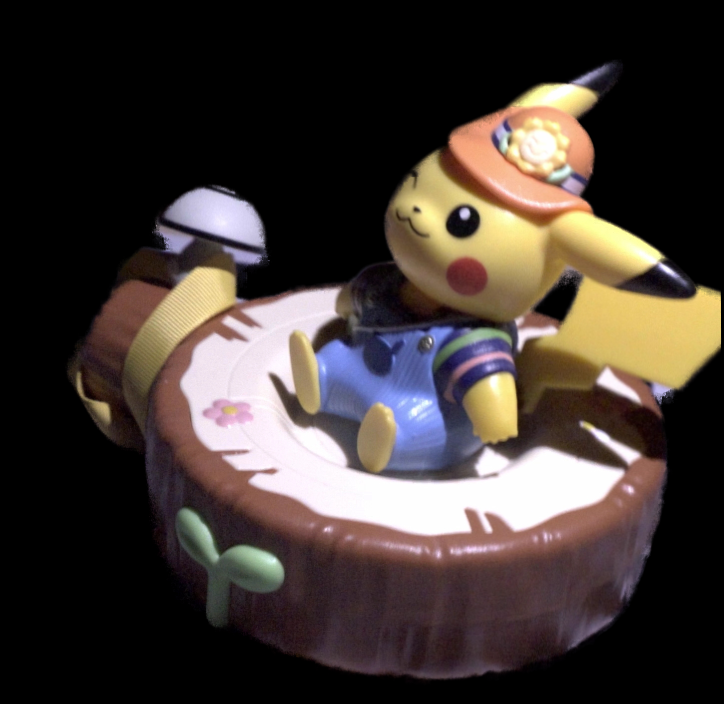}
        \end{minipage}
        \begin{minipage}{1.15in}
            \includegraphics[width=\textwidth]{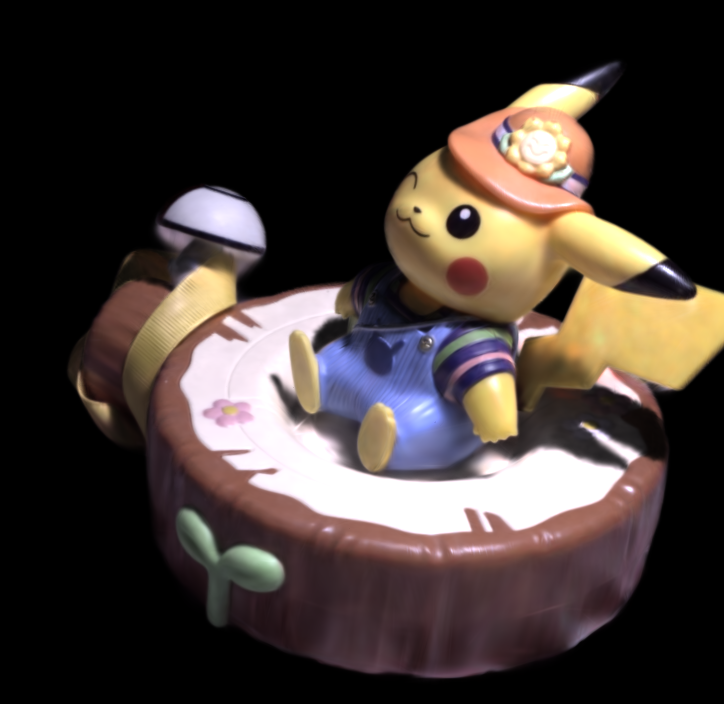}
        \end{minipage}
        \begin{minipage}{1.15in}
            \includegraphics[width=\textwidth]{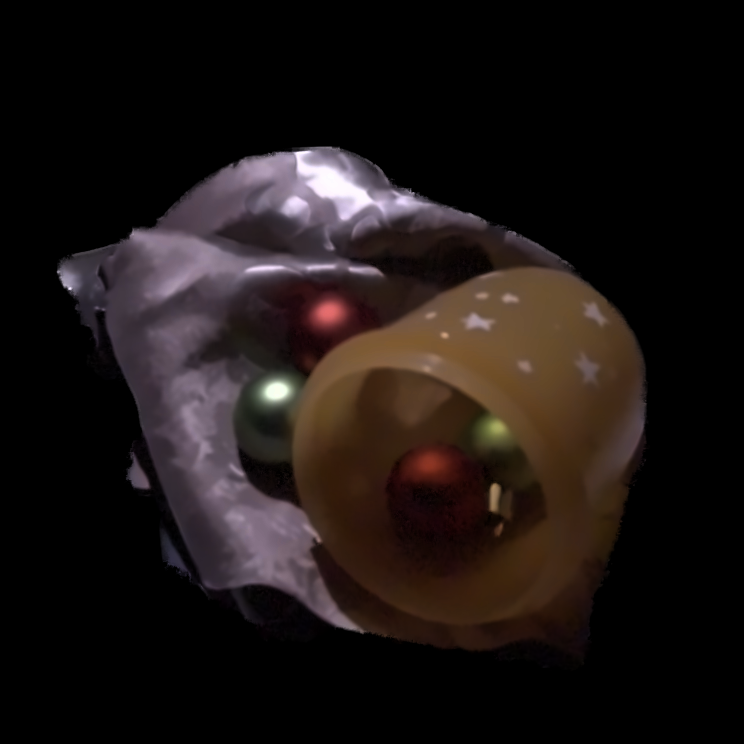}
        \end{minipage}
        \begin{minipage}{1.15in}
            \includegraphics[width=\textwidth]{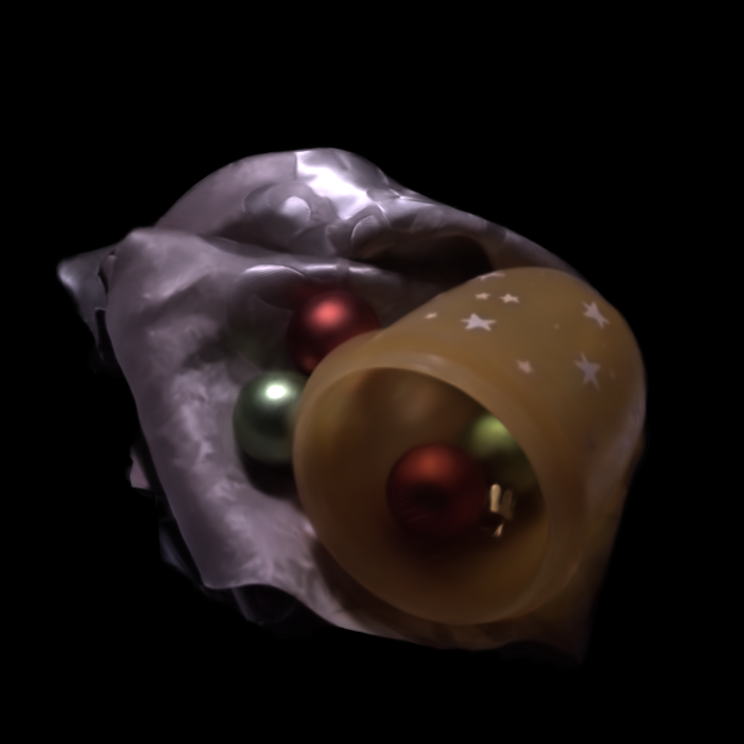}
        \end{minipage}
        \begin{minipage}{1.15in}
            \includegraphics[width=\textwidth]{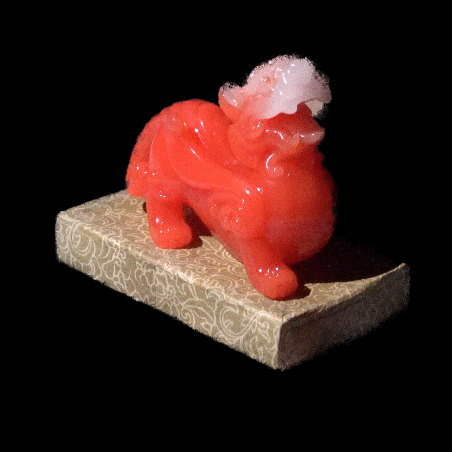}
        \end{minipage}
        \begin{minipage}{1.15in}
            \includegraphics[width=\textwidth]{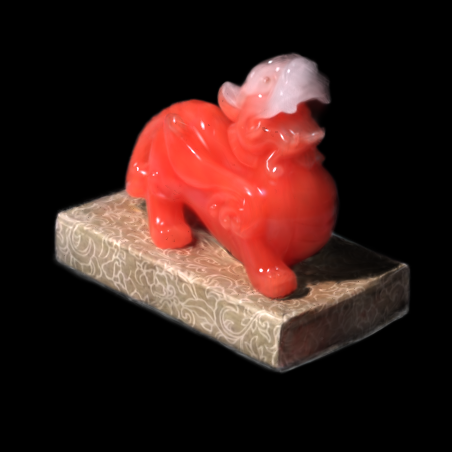}
        \end{minipage}
    \end{minipage}

    \begin{minipage}{\textwidth}
        \centering
        \begin{minipage}{2.2in}
            \centering
            \scalebox{.85}{\sc{Pikachu: 0.9676 | 32.39 | 0.0666}}
        \end{minipage}
        \begin{minipage}{2.2in}
            \centering
            \scalebox{.85}{\sc{Cup-Fabric: 0.9834 | 37.09 | 0.0494}}
        \end{minipage}
        \begin{minipage}{2.2in}
            \centering
            \scalebox{.85}{\sc{Pixiu: 0.9428 | 30.82 | 0.0816}}
        \end{minipage}
    \end{minipage}
    \caption{Our relighting results on captured data from~\cite{zeng2023nrhints}. For each pair of images, the left one is the ground-truth photograph, and the right is our rendering result. Average errors in SSIM, PSNR and LPIPS are reported at the bottom.}
    \label{fig:real evaluation}
\end{figure*}

%% file: figure/figure_only/cmp_OSF.tex
\begin{figure}[htb]

    \begin{minipage}{\linewidth}
        \centering
        \begin{minipage}{0.32\linewidth}
            \centering
            \scalebox{.85}{Ground-Truth}
        \end{minipage}
        \begin{minipage}{0.32\linewidth}
            \centering
            \scalebox{.85}{Ours}
        \end{minipage}
        \begin{minipage}{0.32\linewidth}
            \centering
            \scalebox{.85}{\cite{yu2023osf}}
        \end{minipage}
    \end{minipage}

    \begin{minipage}{\linewidth}
        \centering
        \begin{minipage}{0.32\linewidth}
            \includegraphics[width=\textwidth]{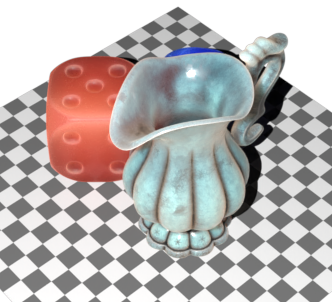}
        \end{minipage}
        \begin{minipage}{0.32\linewidth}
            \includegraphics[width=\textwidth]{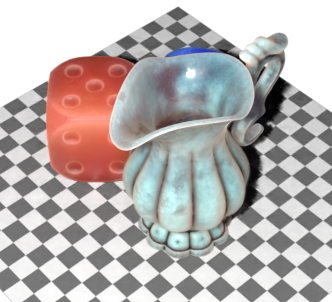}
        \end{minipage}
        \begin{minipage}{0.32\linewidth}
            \includegraphics[width=\textwidth]{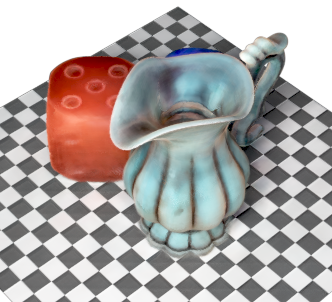}
        \end{minipage}
    \end{minipage}

    \begin{minipage}{\linewidth}
        \centering
        \begin{minipage}{0.32\linewidth}
            \centering
            \scalebox{.85}{\sc{SSIM | PSNR | LPIPS}}
        \end{minipage}
        \begin{minipage}{0.32\linewidth}
            \centering
            \scalebox{.85}{\sc{0.9740 | 32.34 | 0.0318}}
        \end{minipage}
        \begin{minipage}{0.32\linewidth}
            \centering
            \scalebox{.85}{\sc{0.9378 | 26.09 | 0.0508}}
        \end{minipage}
    \end{minipage}

    \caption{Comparison with~\cite{yu2023osf}. From the left to right, the ground-truth, the results of our approach and~\cite{yu2023osf}, respectively. Average errors in SSIM, PSNR and LPIPS are reported at the bottom of related images.}
    
    \label{fig:comparsion_osf}
\end{figure}

%% file: src/RGS-figure-only.tex
\input{figure/figure_only/decompose_vis}

\input{figure/figure_only/cmp_NRHints}

\input{figure/figure_only/cmp_NRTF}

\input{figure/figure_only/eval_syn}

\input{figure/figure_only/cmp_env.tex}

%% file: figure/figure_only/decompose_vis.tex
\begin{figure*}[htb]

    \begin{minipage}{\textwidth}
        \centering
        \begin{minipage}{1.15in}
            \centering
            \scalebox{.85}{Rendering Result}
        \end{minipage}
        \begin{minipage}{1.15in}
            \centering
            \scalebox{.85}{Diffuse Albedo}
        \end{minipage}
        \begin{minipage}{1.15in}
            \centering
            \scalebox{.85}{Specular Albedo}
        \end{minipage}
        \begin{minipage}{1.15in}
            \centering
            \scalebox{.85}{Normal}
        \end{minipage}
        \begin{minipage}{1.15in}
            \centering
            \scalebox{.85}{Shadow Function}
        \end{minipage}
        \begin{minipage}{1.15in}
            \centering
            \scalebox{.85}{Other Effects ($\times2.5$)}
        \end{minipage}
    \end{minipage}
    
    \begin{minipage}{\textwidth}
        \centering
        \begin{minipage}{1.15in}
            \includegraphics[width=\textwidth]{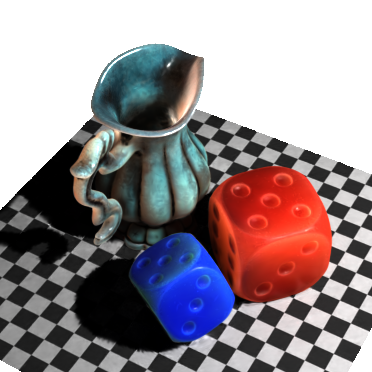}
        \end{minipage}
        \begin{minipage}{1.15in}
            \includegraphics[width=\textwidth]{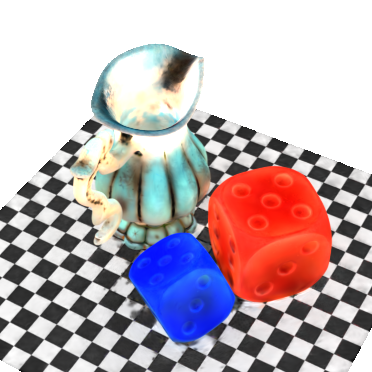}
        \end{minipage}
        \begin{minipage}{1.15in}
            \includegraphics[width=\textwidth]{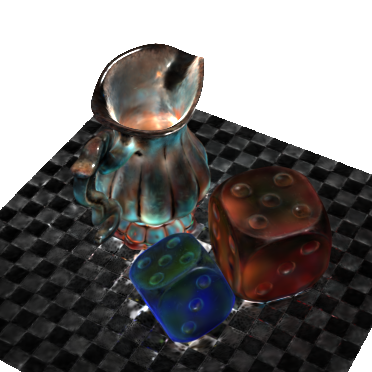}
        \end{minipage}
        \begin{minipage}{1.15in}
            \includegraphics[width=\textwidth]{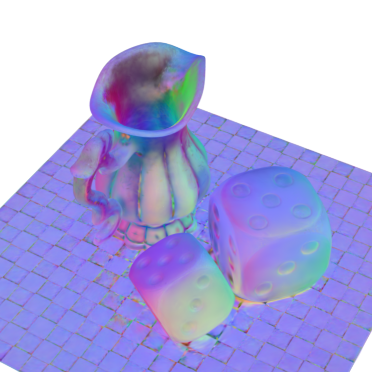}
        \end{minipage}
        \begin{minipage}{1.15in}
            \includegraphics[width=\textwidth]{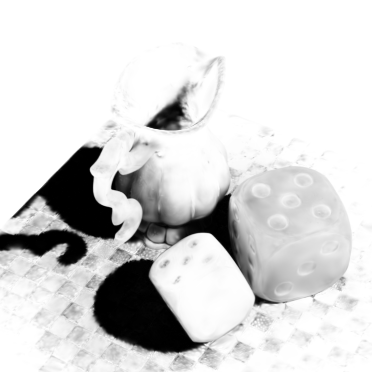}
        \end{minipage}
        \begin{minipage}{1.15in}
            \includegraphics[width=\textwidth]{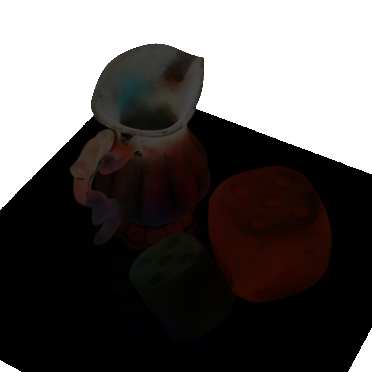}
        \end{minipage}
    \end{minipage}

    \caption{Visualization of various components of our representation. From the left image to right, the final rendering result, diffuse albedo $\mathbf{\rho}_d$, specular albedo $\mathbf{\rho}_s$, normal $\mathbf{n}$, shadow function $\Phi$ and other effects $\Psi$ ($\times2.5$ for a better visualization). An image gamma of 1 is used for this figure.}
    \label{fig:decompose_vis}
\end{figure*}

%% file: figure/figure_only/cmp_NRHints.tex
\begin{figure*}[htb]

    \begin{minipage}{\textwidth}
        \centering
        \begin{minipage}{0.08in}
        \ 
        \end{minipage}
        \begin{minipage}{1.13in}
            \centering
            \scalebox{.85}{\sc{Drums}}
        \end{minipage}
        \begin{minipage}{1.13in}
            \centering
            \scalebox{.85}{\sc{FurBall}}
        \end{minipage}
        \begin{minipage}{1.13in}
            \centering
            \scalebox{.85}{\sc{Lego}}
        \end{minipage}
        \begin{minipage}{1.13in}
            \centering
            \scalebox{.85}{\sc{Fish}}
        \end{minipage}
        \begin{minipage}{1.13in}
            \centering
            \scalebox{.85}{\sc{Cluttered}}
        \end{minipage}
        \begin{minipage}{1.13in}
            \centering
            \scalebox{.85}{\sc{Cat}}
        \end{minipage}
    \end{minipage}

    \begin{minipage}{\textwidth}
        \centering
        \begin{minipage}{0.08in}
            \centering
            \rotatebox{90}{\scalebox{.85}{Ground-Truth}}
        \end{minipage}
        \begin{minipage}{1.13in}
            \begin{minipage}{1.13in}
                \includegraphics[width=\textwidth]{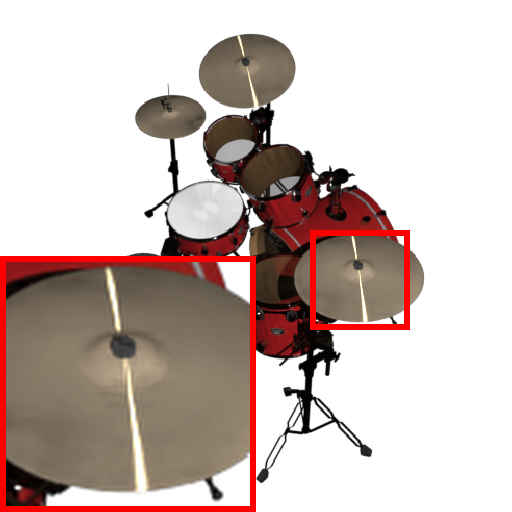}
            \end{minipage}
        \end{minipage}
        \begin{minipage}{1.13in}
           \begin{minipage}{1.13in}
                \includegraphics[width=\textwidth]{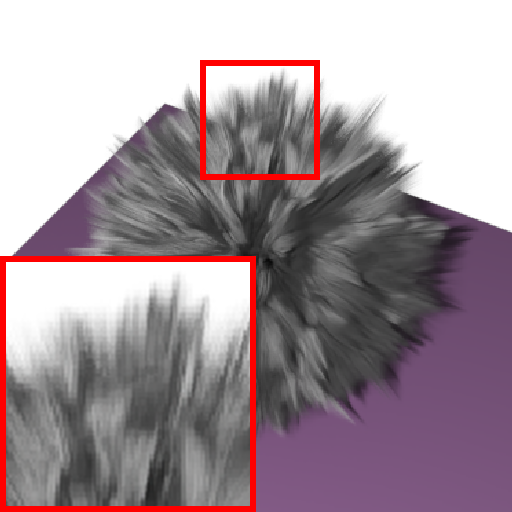}
            \end{minipage}
        \end{minipage}
        \begin{minipage}{1.13in}
            \begin{minipage}{1.13in}
                \includegraphics[width=\textwidth]{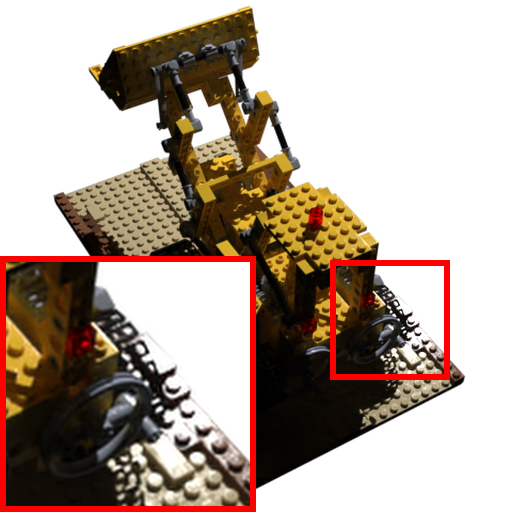}
            \end{minipage}
        \end{minipage}
        \begin{minipage}{1.13in}
           \begin{minipage}{1.13in}
                \includegraphics[width=\textwidth]{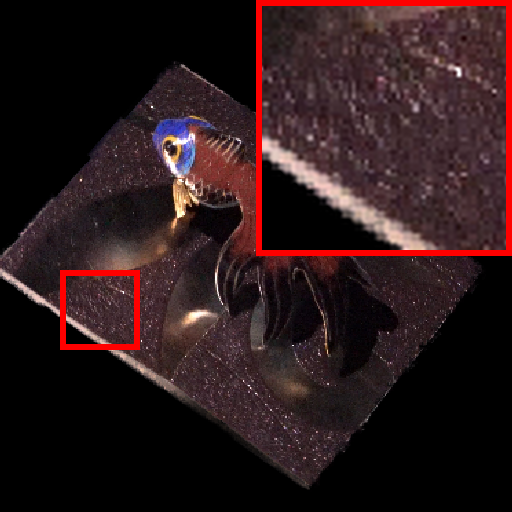}
            \end{minipage}
        \end{minipage}
        \begin{minipage}{1.13in}
            \begin{minipage}{1.13in}
                \includegraphics[width=\textwidth]{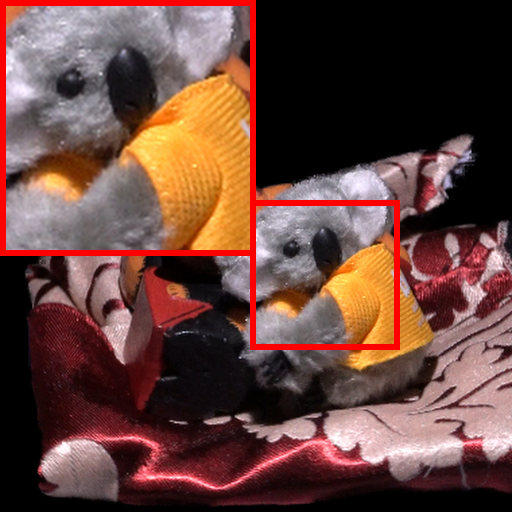}
            \end{minipage}
        \end{minipage}
        \begin{minipage}{1.13in}
            \begin{minipage}{1.13in}
                \includegraphics[width=\textwidth]{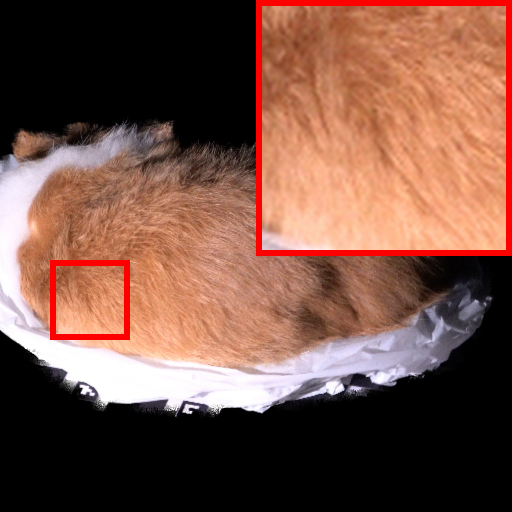}
            \end{minipage}
        \end{minipage}
    \end{minipage}

    \begin{minipage}{\textwidth}
        \centering
        \begin{minipage}{0.08in}
            \centering
            \rotatebox{90}{\scalebox{.85}{Ours}}
        \end{minipage}
        \begin{minipage}{1.13in}
            \begin{minipage}{1.13in}
                \includegraphics[width=\textwidth]{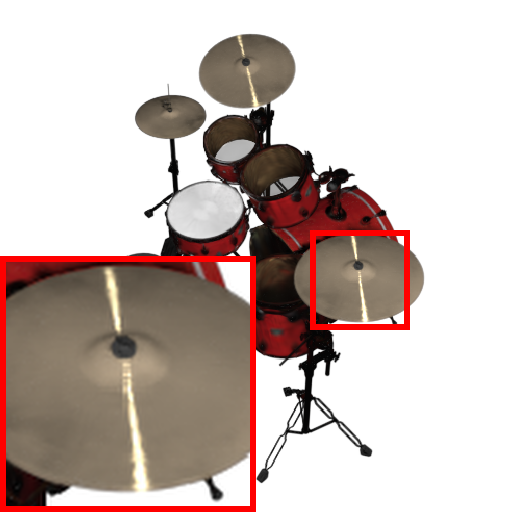}
            \end{minipage}
        \end{minipage}
        \begin{minipage}{1.13in}
           \begin{minipage}{1.13in}
                \includegraphics[width=\textwidth]{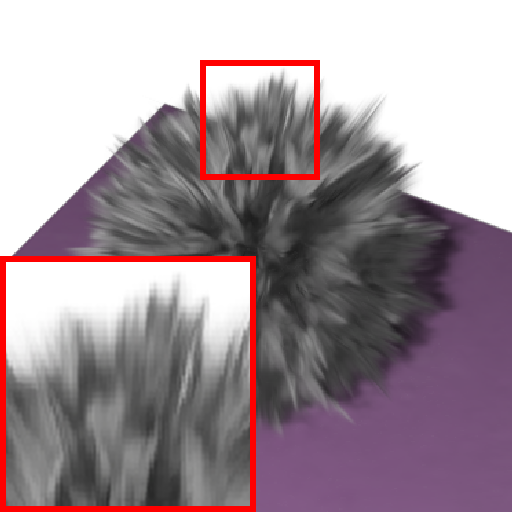}
            \end{minipage}
        \end{minipage}
        \begin{minipage}{1.13in}
            \begin{minipage}{1.13in}
                \includegraphics[width=\textwidth]{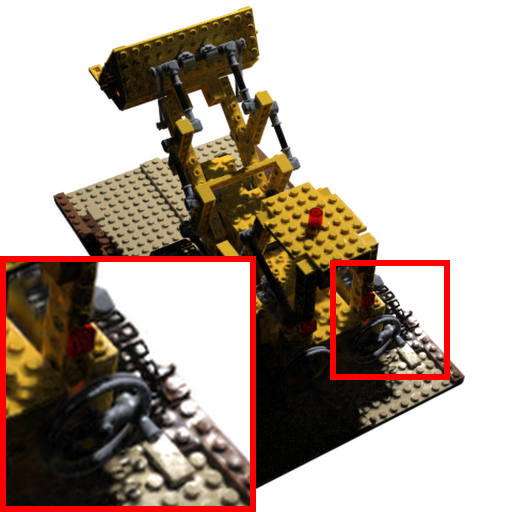}
            \end{minipage}
        \end{minipage}
        \begin{minipage}{1.13in}
            \begin{minipage}{1.13in}
                \includegraphics[width=\textwidth]{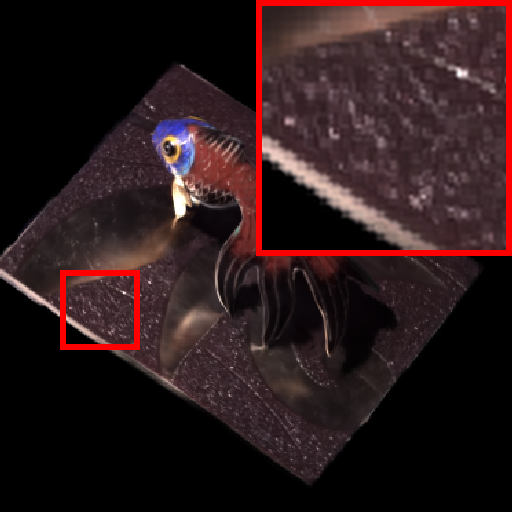}
            \end{minipage}
        \end{minipage}
        \begin{minipage}{1.13in}
            \begin{minipage}{1.13in}
                \includegraphics[width=\textwidth]{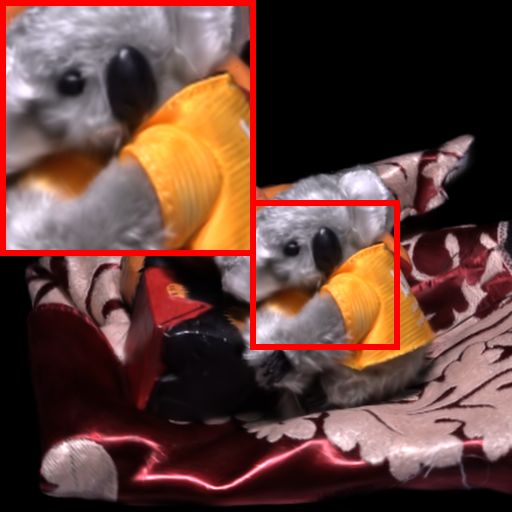}
            \end{minipage}
        \end{minipage}
        \begin{minipage}{1.13in}
            \begin{minipage}{1.13in}
                \includegraphics[width=\textwidth]{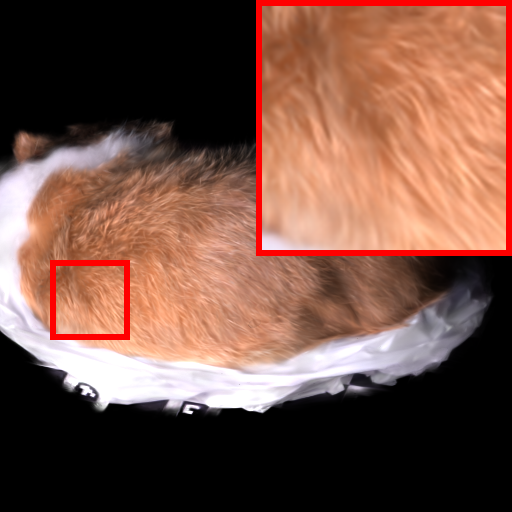}
            \end{minipage}
        \end{minipage}
    \end{minipage}

    \begin{minipage}{\textwidth}
        \centering
        \begin{minipage}{0.08in}
        \ 
        \end{minipage}
        \begin{minipage}{1.13in}
            \centering
            \scalebox{.85}{\sc{0.9714 |30.45 | 0.0276}}
        \end{minipage}
        \begin{minipage}{1.13in}
            \centering
            \scalebox{.85}{\sc{0.9669 | 35.34 | 0.0534}}
        \end{minipage}
        \begin{minipage}{1.13in}
            \centering
            \scalebox{.85}{\sc{0.9511 | 30.19 | 0.0468}}
        \end{minipage}
        \begin{minipage}{1.13in}
            \centering
            \scalebox{.85}{\sc{0.9252 | 31.13 | 0.0866}}
        \end{minipage}
        \begin{minipage}{1.13in}
            \centering
            \scalebox{.85}{\sc{0.9521 | 30.82 | 0.0696}}
        \end{minipage}
        \begin{minipage}{1.13in}
            \centering
            \scalebox{.85}{\sc{0.8981 | 26.43 | 0.1381}}
        \end{minipage}
    \end{minipage}

    \begin{minipage}{\textwidth}
        \centering
        \begin{minipage}{0.08in}
            \centering
            \rotatebox{90}{\scalebox{.85}{NRHints~\cite{zeng2023nrhints}}}
        \end{minipage}
        \begin{minipage}{1.13in}
            \begin{minipage}{1.13in}
                \includegraphics[width=\textwidth]{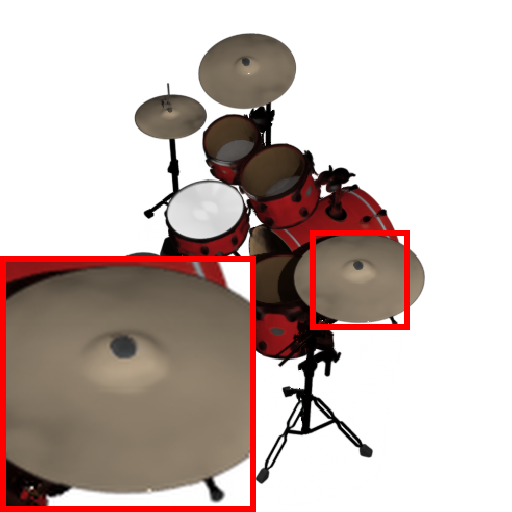}
            \end{minipage}
        \end{minipage}
        \begin{minipage}{1.13in}
           \begin{minipage}{1.13in}
                \includegraphics[width=\textwidth]{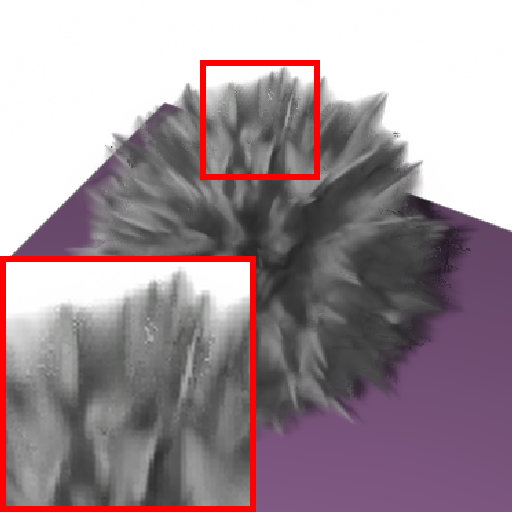}
            \end{minipage}
        \end{minipage}
        \begin{minipage}{1.13in}
            \begin{minipage}{1.13in}
                \includegraphics[width=\textwidth]{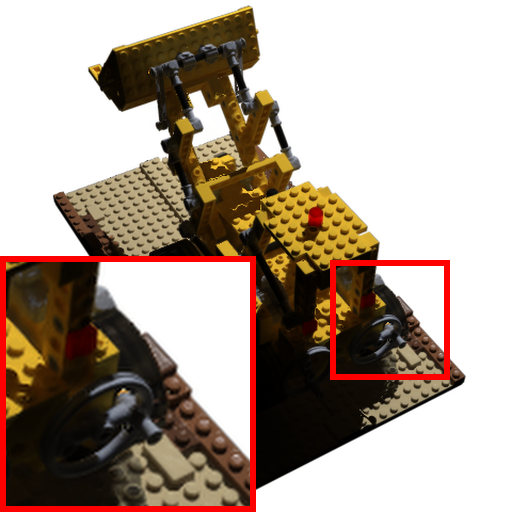}
            \end{minipage}
        \end{minipage}
        \begin{minipage}{1.13in}
            \begin{minipage}{1.13in}
                \includegraphics[width=\textwidth]{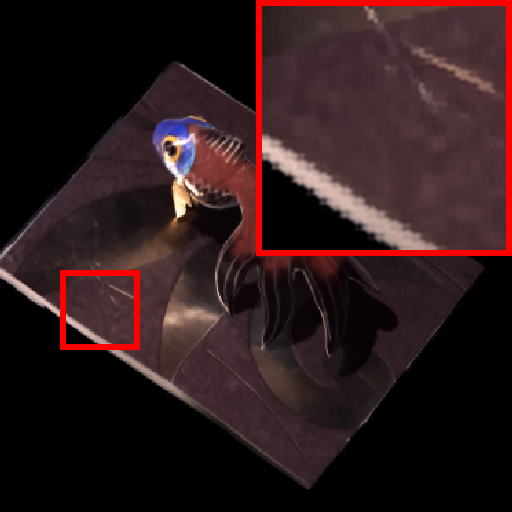}
            \end{minipage}
        \end{minipage}
        \begin{minipage}{1.13in}
            \begin{minipage}{1.13in}
                \includegraphics[width=\textwidth]{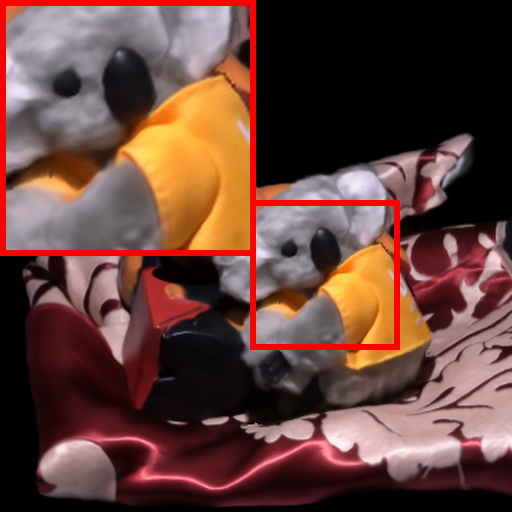}
            \end{minipage}
        \end{minipage}
        \begin{minipage}{1.13in}
            \begin{minipage}{1.13in}
                \includegraphics[width=\textwidth]{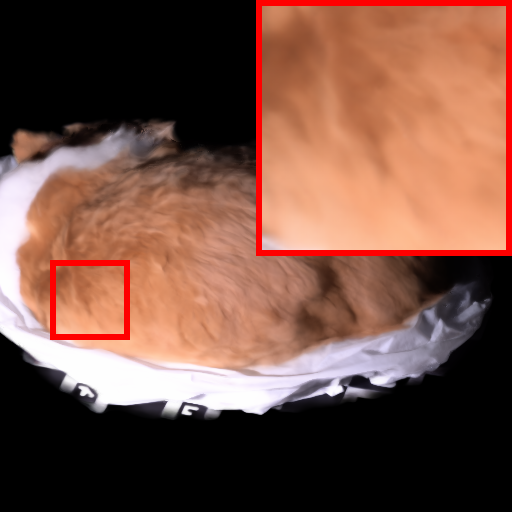}
            \end{minipage}
        \end{minipage}
    \end{minipage}

    \begin{minipage}{\textwidth}
        \centering
        \begin{minipage}{0.08in}
        \ 
        \end{minipage}
        \begin{minipage}{1.13in}
            \centering
            \scalebox{.85}{\sc{0.9745 | 29.88 | 0.0294}}
        \end{minipage}
        \begin{minipage}{1.13in}
            \centering
            \scalebox{.85}{\sc{0.9522 | 35.06 | 0.0777}}
        \end{minipage}
        \begin{minipage}{1.13in}
            \centering
            \scalebox{.85}{\sc{0.9583 | 29.90 | 0.0393}}
        \end{minipage}
        \begin{minipage}{1.13in}
            \centering
            \scalebox{.85}{\sc{0.9140 | 31.28 | 0.1137}}
        \end{minipage}
        \begin{minipage}{1.13in}
            \centering
            \scalebox{.85}{\sc{0.9280 | 29.61 | 0.0879}}
        \end{minipage}
        \begin{minipage}{1.13in}
            \centering
            \scalebox{.85}{\sc{0.8560 | 27.34 | 0.1667}}
        \end{minipage}
    \end{minipage}
    
    \caption{Comparisons to~\cite{zeng2023nrhints}. From the top row to bottom, the ground-truth, results of our approach and~\cite{zeng2023nrhints}, respectively. From the 1st column to 3rd, synthetic data from \cite{mildenhall2020nerf}; from the 4th column to the last, captured data from \cite{zeng2023nrhints}. Average errors in SSIM, PSNR and LPIPS are reported at the bottom of each related image.}
    \label{fig:pointlight_comparison}
\end{figure*}

%% file: figure/figure_only/cmp_NRTF.tex
\begin{figure*}[htb]

    \begin{minipage}{\textwidth}
        \centering
        \begin{minipage}{1.15in}
            \centering
            \scalebox{.85}{Ground-Truth}
        \end{minipage}
        \begin{minipage}{1.15in}
            \centering
            \scalebox{.85}{Ours}
        \end{minipage}
        \begin{minipage}{1.15in}
            \centering
            \scalebox{.85}{NRTF~\cite{lyu2022neural}}
        \end{minipage}
        \begin{minipage}{1.15in}
            \centering
            \scalebox{.85}{Ground-Truth}
        \end{minipage}
        \begin{minipage}{1.15in}
            \centering
            \scalebox{.85}{Ours}
        \end{minipage}
        \begin{minipage}{1.15in}
            \centering
            \scalebox{.85}{NRTF~\cite{lyu2022neural}}
        \end{minipage}
    \end{minipage}

    \begin{minipage}{\textwidth}
        \centering
        \begin{minipage}{1.15in}
           \begin{minipage}{1.15in}
                \includegraphics[width=\textwidth]{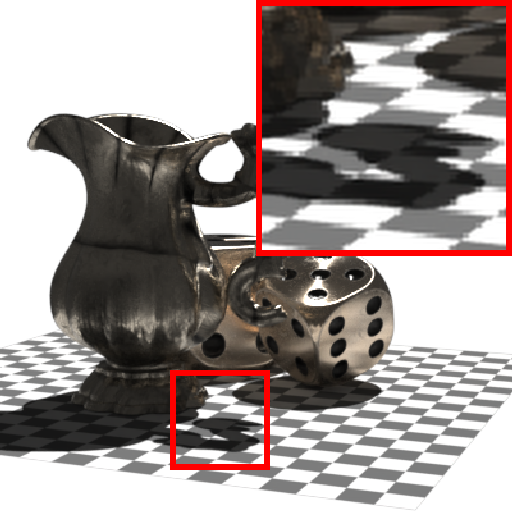}
            \end{minipage}
        \end{minipage}
        \begin{minipage}{1.15in}
            \begin{minipage}{1.15in}
                \includegraphics[width=\textwidth]{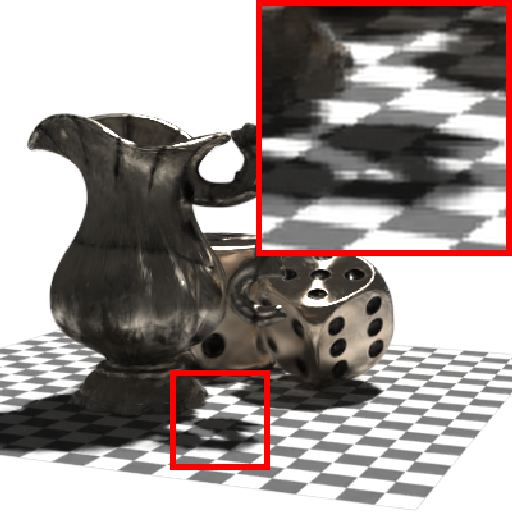}
            \end{minipage}
        \end{minipage}
        \begin{minipage}{1.15in}
            \begin{minipage}{1.15in}
                \includegraphics[width=\textwidth]{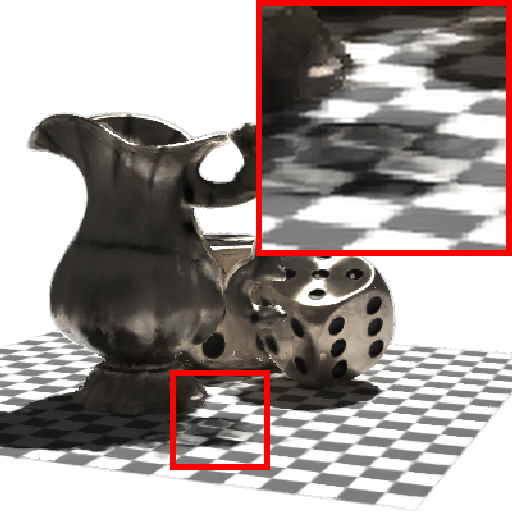}
            \end{minipage}
        \end{minipage}
        \begin{minipage}{1.15in}
           \begin{minipage}{1.15in}
                \includegraphics[width=\textwidth]{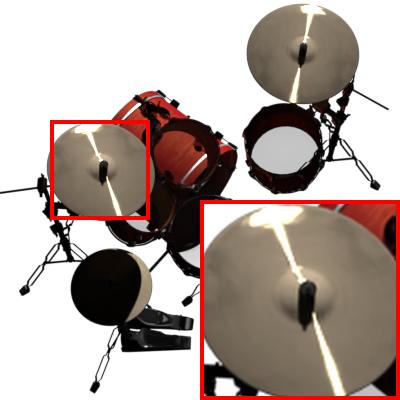}
            \end{minipage}
        \end{minipage}
        \begin{minipage}{1.15in}
            \begin{minipage}{1.15in}
                \includegraphics[width=\textwidth]{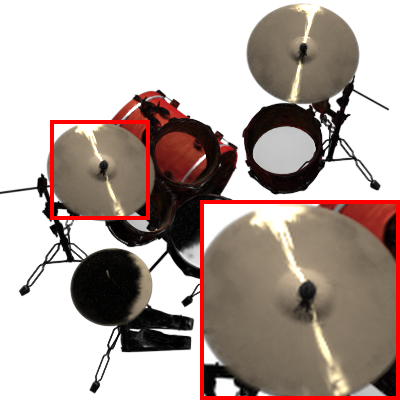}
            \end{minipage}
        \end{minipage}
        \begin{minipage}{1.15in}
            \begin{minipage}{1.15in}
                \includegraphics[width=\textwidth]{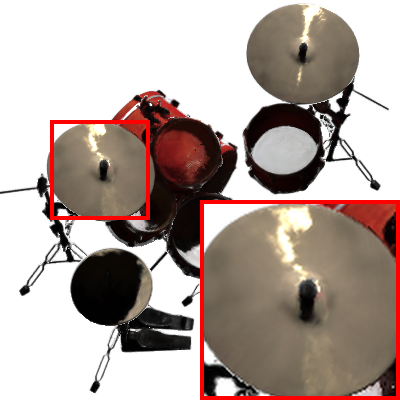}
            \end{minipage}
        \end{minipage}
    \end{minipage}
    
    \begin{minipage}{\textwidth}
        \centering
        \begin{minipage}{1.15in}
            \centering
            \scalebox{.85}{\sc{SSIM | PSNR | LPIPS}}
        \end{minipage}
        \begin{minipage}{1.15in}
            \centering
            \scalebox{.85}{\sc{0.9568 | 27.83 | 0.0399 }}
        \end{minipage}
        \begin{minipage}{1.15in}
            \centering
            \scalebox{.85}{\sc{0.9494 | 26.89 | 0.0486 }}
        \end{minipage}
        \begin{minipage}{1.15in}
            \centering
            \scalebox{.85}{\sc{SSIM | PSNR | LPIPS}}
        \end{minipage}
        \begin{minipage}{1.15in}
            \centering
            \scalebox{.85}{\sc{0.9686 | 30.43 | 0.0275 }}
        \end{minipage}
        \begin{minipage}{1.15in}
            \centering
            \scalebox{.85}{\sc{0.9606 | 27.09 | 0.0391 }}
        \end{minipage}
    \end{minipage}

    \caption{Comparisons to~\cite{lyu2022neural}. From every 3 consecutive images, the ground-truth, result with our approach and~\cite{lyu2022neural}, respectively. Average errors in SSIM, PSNR and LPIPS are reported at the bottom of each related image.}
    \label{fig:NRTF_comparison}
\end{figure*}

%% file: figure/figure_only/eval_syn.tex
\begin{figure*}[htb]

    \begin{minipage}{\textwidth}
        \centering
        \begin{minipage}{1.15in}
            \centering
            \scalebox{.85}{Ground-Truth}
        \end{minipage}
        \begin{minipage}{1.15in}
            \centering
            \scalebox{.85}{Ours}
        \end{minipage}
        \begin{minipage}{1.15in}
            \centering
            \scalebox{.85}{Ground-Truth}
        \end{minipage}
        \begin{minipage}{1.15in}
            \centering
            \scalebox{.85}{Ours}
        \end{minipage}
        \begin{minipage}{1.15in}
            \centering
            \scalebox{.85}{Ground-Truth}
        \end{minipage}
        \begin{minipage}{1.15in}
            \centering
            \scalebox{.85}{Ours}
        \end{minipage}
    \end{minipage}

    \begin{minipage}{\textwidth}
        \centering
        \begin{minipage}{1.15in}
            \includegraphics[width=\textwidth]{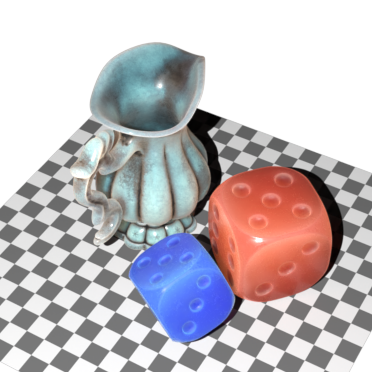}
        \end{minipage}
        \begin{minipage}{1.15in}
            \includegraphics[width=\textwidth]{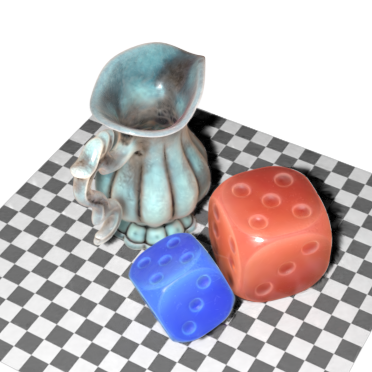}
        \end{minipage}
        \begin{minipage}{1.15in}
            \includegraphics[width=\textwidth]{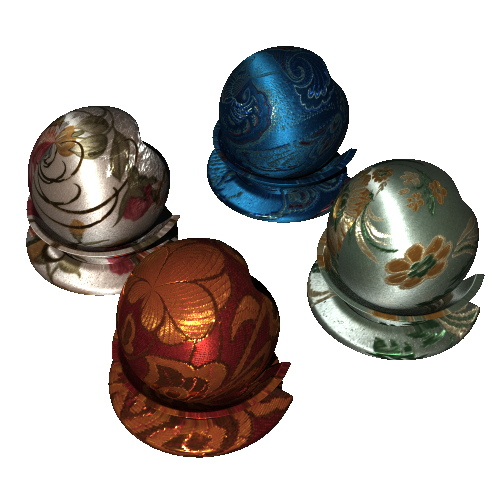}
        \end{minipage}
        \begin{minipage}{1.15in}
            \includegraphics[width=\textwidth]{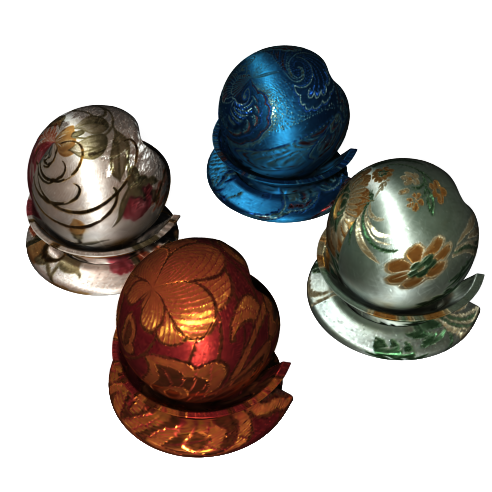}
        \end{minipage}
        \begin{minipage}{1.15in}
            \includegraphics[width=\textwidth]{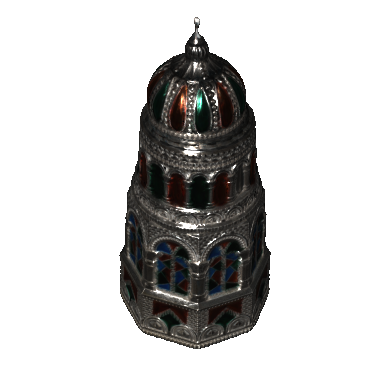}
        \end{minipage}
        \begin{minipage}{1.15in}
            \includegraphics[width=\textwidth]{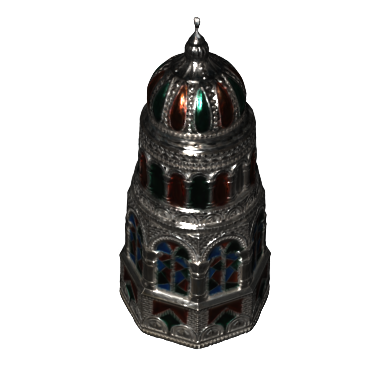}
        \end{minipage}
    \end{minipage}
    \begin{minipage}{\textwidth}
        \centering
        \begin{minipage}{2.2in}
            \centering
            \scalebox{.85}{\sc{Translucent: 0.9740 | 32.34 | 0.0318}}
        \end{minipage}
        \begin{minipage}{2.2in}
            \centering
            \scalebox{.85}{\sc{MaterialBalls: 0.9545 | 28.04 | 0.0628}}
        \end{minipage}
        \begin{minipage}{2.2in}
            \centering
            \scalebox{.85}{\sc{Tower: 0.9913 | 32.96 | 0.0150}}
        \end{minipage}
    \end{minipage}

    \begin{minipage}{\textwidth}
        \centering
        \begin{minipage}{1.15in}
            \includegraphics[width=\textwidth]{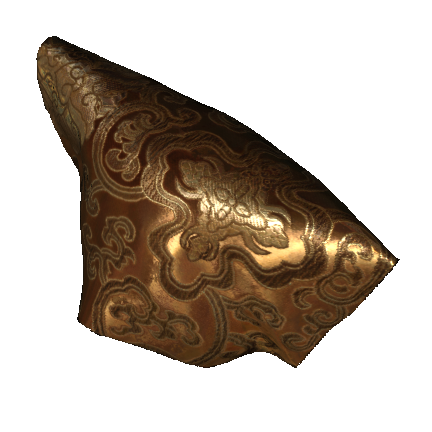}
        \end{minipage}
        \begin{minipage}{1.15in}
            \includegraphics[width=\textwidth]{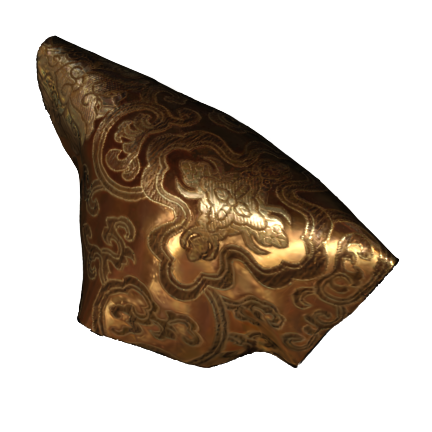}
        \end{minipage}
        \begin{minipage}{1.15in}
            \includegraphics[width=\textwidth]{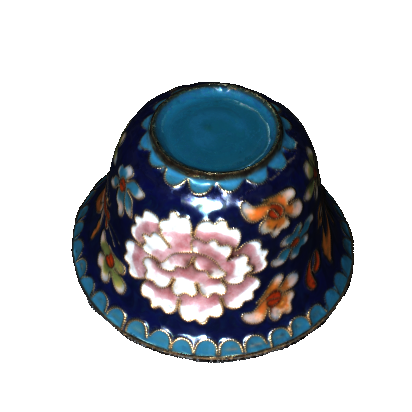}
        \end{minipage}
        \begin{minipage}{1.15in}
            \includegraphics[width=\textwidth]{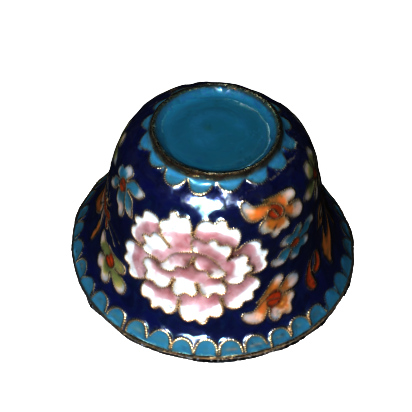}
        \end{minipage}
        \begin{minipage}{1.15in}
            \includegraphics[width=\textwidth]{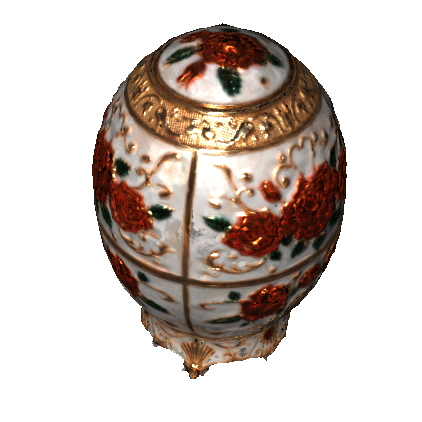}
        \end{minipage}
        \begin{minipage}{1.15in}
            \includegraphics[width=\textwidth]{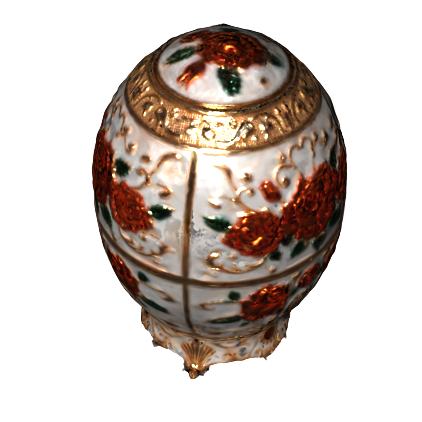}
        \end{minipage}
    \end{minipage}

    \begin{minipage}{\textwidth}
        \centering
        \begin{minipage}{2.2in}
            \centering
            \scalebox{.85}{\sc{Fabric: 0.9812 | 31.15 | 0.0317}}
        \end{minipage}
        \begin{minipage}{2.2in}
            \centering
            \scalebox{.85}{\sc{Cup: 0.9905 | 32.74 | 0.0203}}
        \end{minipage}
        \begin{minipage}{2.2in}
            \centering
            \scalebox{.85}{\sc{Egg: 0.9833 | 31.84 | 0.0257}}
        \end{minipage}
    \end{minipage}
    
    \caption{Our relighting results on synthetic data/rendered images of captured data~\cite{zeng2023nrhints, kang2019learning, ma2023open}. For each pair of images, the left one is the ground-truth, and the right is our result. Average errors in SSIM, PSNR and LPIPS are reported at the bottom.}
    \label{fig:synthetic-evaluation}
\end{figure*}

%% file: figure/figure_only/cmp_env.tex
\begin{figure*}[htb]

    \begin{minipage}{\textwidth}
        \centering
        \begin{minipage}{0.08in}
        \ 
        \end{minipage}
        \begin{minipage}{1.13in}
            \centering
            \scalebox{.85}{Ground-Truth}
        \end{minipage}
        \begin{minipage}{1.13in}
            \centering
            \scalebox{.85}{Ours}
        \end{minipage}
        \begin{minipage}{1.13in}
            \centering
            \scalebox{.85}{GaussianShader}
        \end{minipage}
        \begin{minipage}{1.13in}
            \centering
            \scalebox{.85}{GS-IR}
        \end{minipage}
        \begin{minipage}{1.13in}
            \centering
            \scalebox{.85}{Relightable 3D Gaussian}
        \end{minipage}
        \begin{minipage}{1.13in}
            \centering
            \scalebox{.85}{TensoIR}
        \end{minipage}
    \end{minipage}

    \begin{minipage}{\textwidth}
        \centering
        \begin{minipage}{0.08in}
            \rotatebox{90}{\scalebox{.85}{\sc{Hotdog}}}
        \end{minipage}
        \begin{minipage}{1.13in}
            \includegraphics[width=\textwidth]{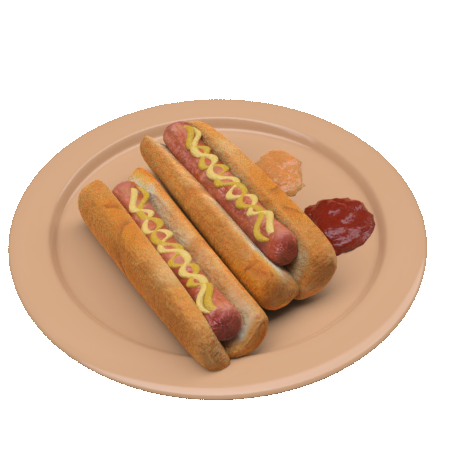}
        \end{minipage}
        \begin{minipage}{1.13in}
            \includegraphics[width=\textwidth]{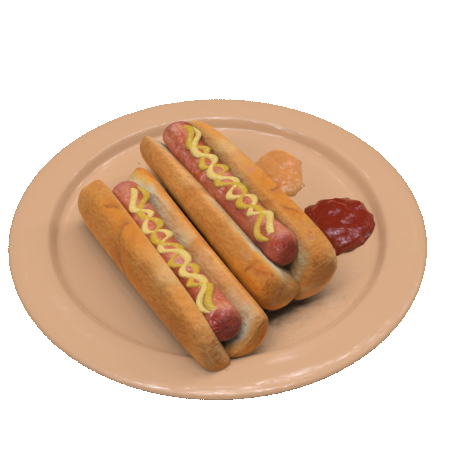}
        \end{minipage}
        \begin{minipage}{1.13in}
            \includegraphics[width=\textwidth]{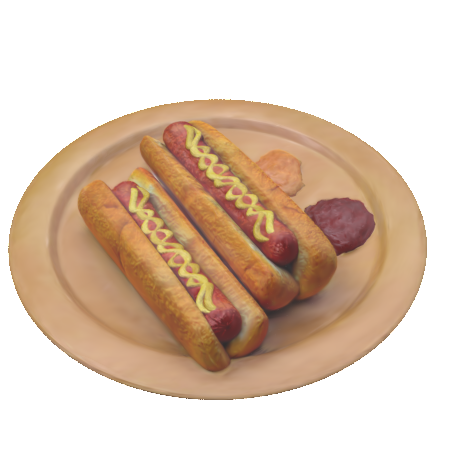}
        \end{minipage}
        \begin{minipage}{1.13in}
            \includegraphics[width=\textwidth]{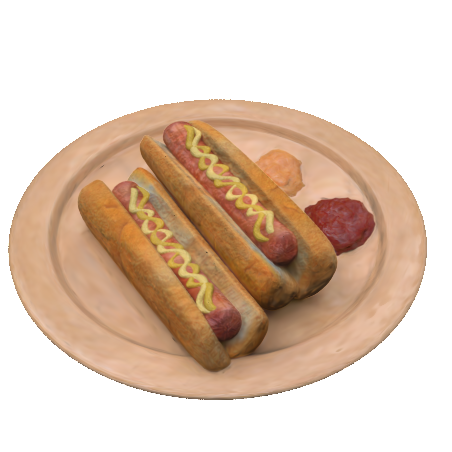}
        \end{minipage}
        \begin{minipage}{1.13in}
            \includegraphics[width=\textwidth]{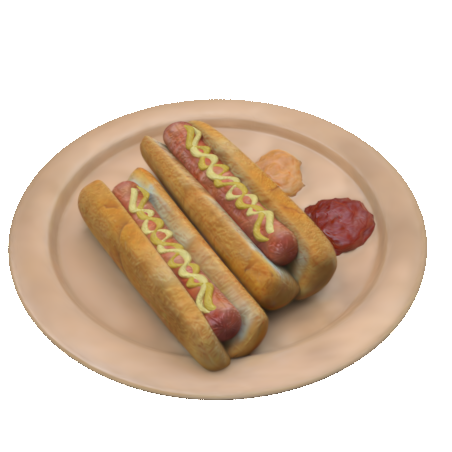}
        \end{minipage}
        \begin{minipage}{1.13in}
            \includegraphics[width=\textwidth]{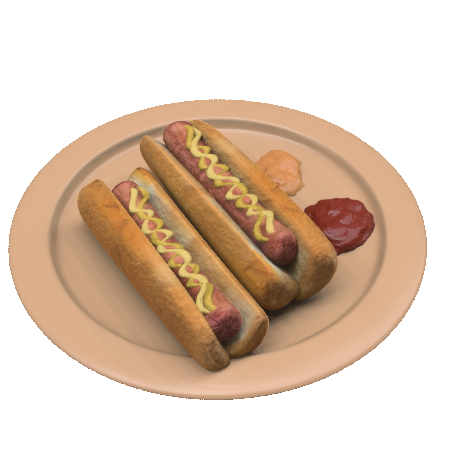}
        \end{minipage}
    \end{minipage}

    \begin{minipage}{\textwidth}
        \centering
        \begin{minipage}{0.08in}
        \ 
        \end{minipage}
        \begin{minipage}{1.13in}
            \centering
            \scalebox{.85}{SSIM | PSNR | LPIPS}
        \end{minipage}
        \begin{minipage}{1.13in}
            \centering
            \scalebox{.85}{0.9712 | 36.47 | 0.0342}
        \end{minipage}
        \begin{minipage}{1.13in}
            \centering
            \scalebox{.85}{0.9439 | 29.25 | 0.0600}
        \end{minipage}
        \begin{minipage}{1.13in}
            \centering
            \scalebox{.85}{0.9289 | 29.13 | 0.0816}
        \end{minipage}
        \begin{minipage}{1.13in}
            \centering
            \scalebox{.85}{0.9526 | 30.22 | 0.0555}
        \end{minipage}
        \begin{minipage}{1.13in}
            \centering
            \scalebox{.85}{0.9590 | 31.68 | 0.0466}
        \end{minipage}
    \end{minipage}

    \begin{minipage}{\textwidth}
        \centering
        \begin{minipage}{0.06in}
            \rotatebox{90}{\scalebox{.85}{\sc{Lego}}}
        \end{minipage}
        \begin{minipage}{1.13in}
            \includegraphics[width=\textwidth]{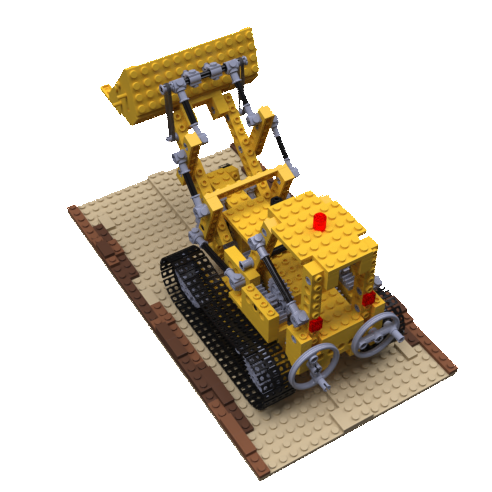}
        \end{minipage}
        \begin{minipage}{1.13in}
            \includegraphics[width=\textwidth]{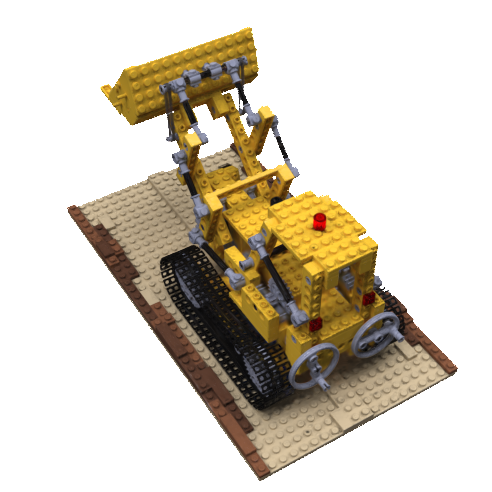}
        \end{minipage}
        \begin{minipage}{1.13in}
            \includegraphics[width=\textwidth]{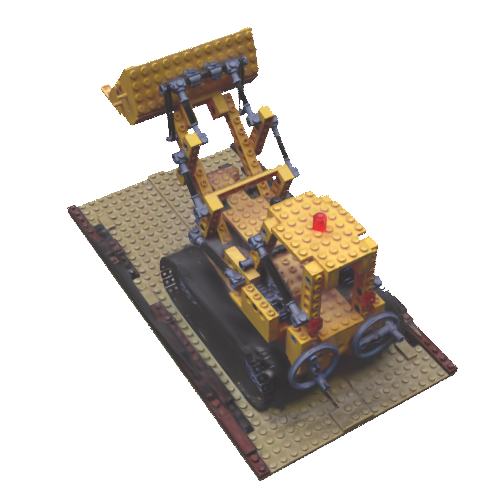}
        \end{minipage}
        \begin{minipage}{1.13in}
            \includegraphics[width=\textwidth]{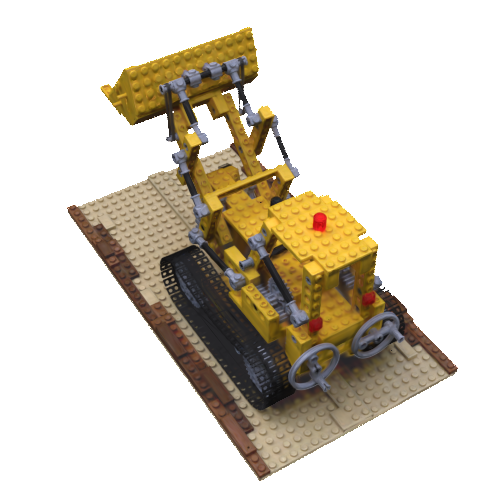}
        \end{minipage}
        \begin{minipage}{1.13in}
            \includegraphics[width=\textwidth]{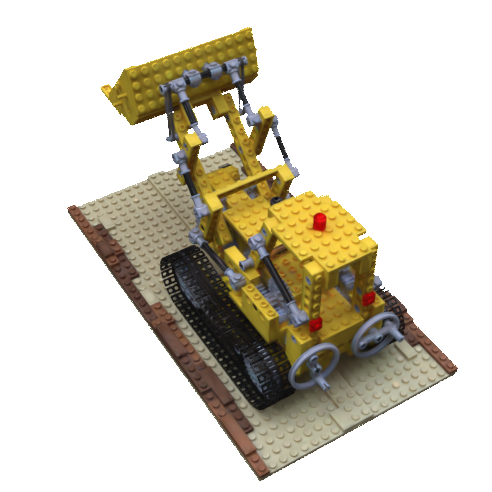}
        \end{minipage}
        \begin{minipage}{1.13in}
            \includegraphics[width=\textwidth]{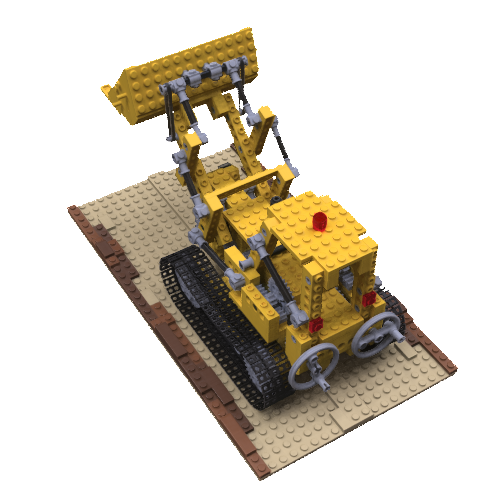}
        \end{minipage}
    \end{minipage}

    \begin{minipage}{\textwidth}
        \centering
        \begin{minipage}{0.08in}
        \ 
        \end{minipage}
        \begin{minipage}{1.13in}
            \centering
            \scalebox{.85}{SSIM | PSNR | LPIPS}
        \end{minipage}
        \begin{minipage}{1.13in}
            \centering
            \scalebox{.85}{0.9651 | 32.06 | 0.0302}
        \end{minipage}
        \begin{minipage}{1.13in}
            \centering
            \scalebox{.85}{0.8964 | 24.33 | 0.0834}
        \end{minipage}
        \begin{minipage}{1.13in}
            \centering
            \scalebox{.85}{0.9230 | 27.66 | 0.0550}
        \end{minipage}
        \begin{minipage}{1.13in}
            \centering
            \scalebox{.85}{0.9463 | 30.31 | 0.0446}
        \end{minipage}
        \begin{minipage}{1.13in}
            \centering
            \scalebox{.85}{0.9540 | 30.96 | 0.0356}
        \end{minipage}
    \end{minipage}

    \begin{minipage}{\textwidth}
        \centering
        \begin{minipage}{0.06in}
            \rotatebox{90}{\scalebox{.85}{\sc{MaterialBalls}}}
        \end{minipage}
        \begin{minipage}{1.13in}
            \includegraphics[width=\textwidth]{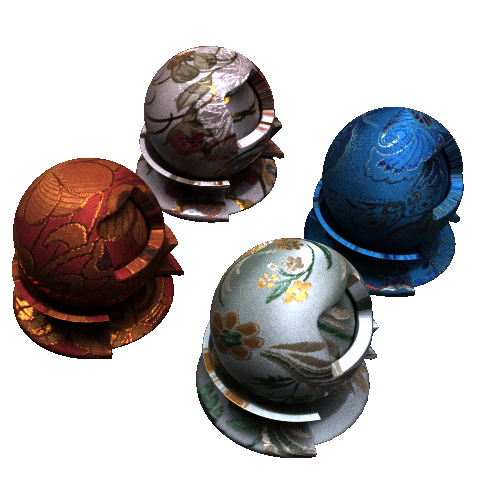}
        \end{minipage}
        \begin{minipage}{1.13in}
            \includegraphics[width=\textwidth]{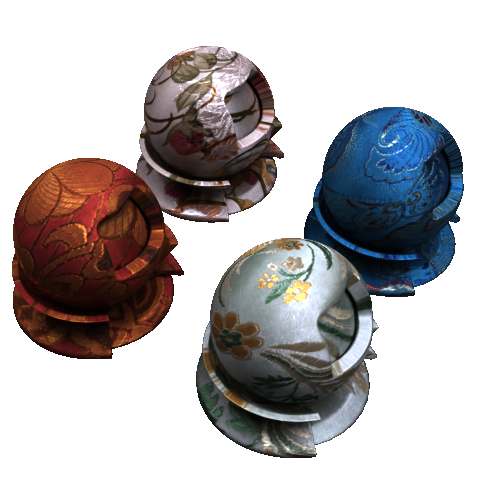}
        \end{minipage}
        \begin{minipage}{1.13in}
            \includegraphics[width=\textwidth]{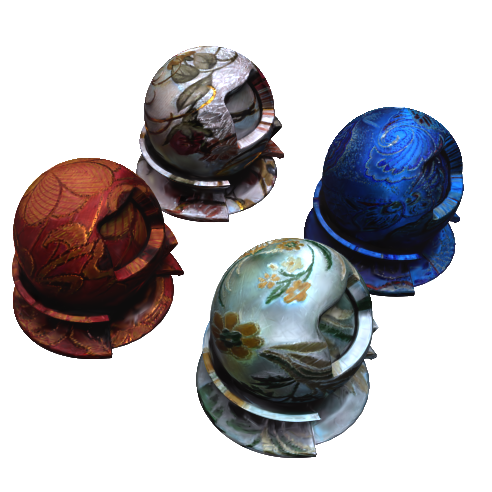}
        \end{minipage}
        \begin{minipage}{1.13in}
            \includegraphics[width=\textwidth]{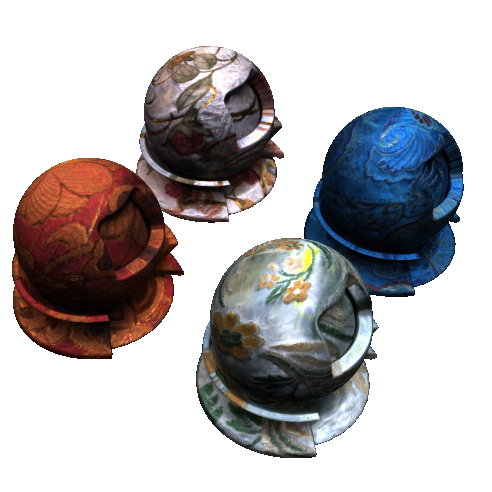}
        \end{minipage}
        \begin{minipage}{1.13in}
            \includegraphics[width=\textwidth]{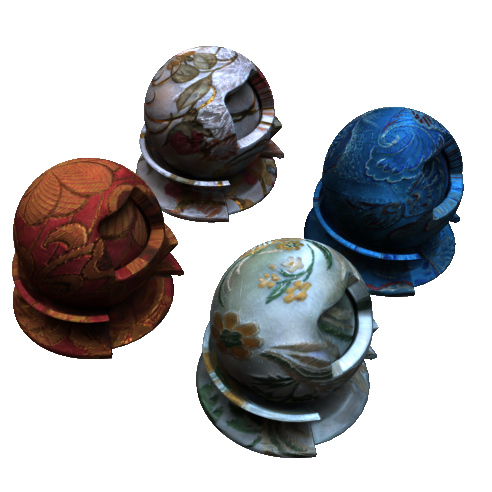}
        \end{minipage}
        \begin{minipage}{1.13in}
            \includegraphics[width=\textwidth]{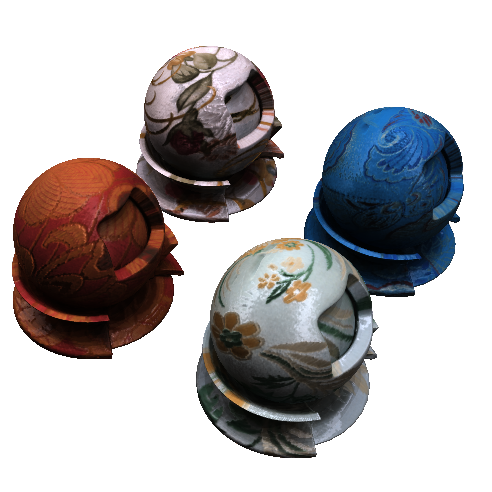}
        \end{minipage}
    \end{minipage}

    \begin{minipage}{\textwidth}
        \centering
        \begin{minipage}{0.08in}
        \ 
        \end{minipage}
        \begin{minipage}{1.13in}
            \centering
            \scalebox{.85}{SSIM | PSNR | LPIPS}
        \end{minipage}
        \begin{minipage}{1.13in}
            \centering
            \scalebox{.85}{0.9459 | 28.01 | 0.0508}
        \end{minipage}
        \begin{minipage}{1.13in}
            \centering
            \scalebox{.85}{0.8900 | 24.31 | 0.0937}
        \end{minipage}
        \begin{minipage}{1.13in}
            \centering
            \scalebox{.85}{0.8756 | 23.41 | 0.0969}
        \end{minipage}
        \begin{minipage}{1.13in}
            \centering
            \scalebox{.85}{0.9045 | 25.34 | 0.0848}
        \end{minipage}
        \begin{minipage}{1.13in}
            \centering
            \scalebox{.85}{0.9026 | 25.05 | 0.0765}
        \end{minipage}
    \end{minipage}

    \caption{Comparisons with approaches using environment-lit input images. For images from the left column to right in each row: the ground-truth, rendering results of our approach, GaussianShader~\cite{jiang2023gaussianshader}, GS-IR~\cite{liang2023gsir}, Relightable 3D Gaussian~\cite{gao2023relightable} and TensoIR~\cite{jin2023tensoir}, respectively. Average errors in SSIM, PSNR and LPIPS are reported at the bottom of each related image.}
    \label{fig:envlight_comparison}
\end{figure*}

%% file: src/supp_body.tex
\section{Ablations}

\input{figure/figure_only/ab_asgnum}

\input{figure/figure_only/ab_photonum}

\subsection{Angular Gaussian vs. MLP} 
In~\figref{fig:ablation}, we replace the appearance function with an MLP of \{30, 100, 600, 100, 30\}. The input of the MLP are $\boldsymbol{\omega}_i, \boldsymbol{\omega}_o$ and a 32D latent vector, and the output is a shading color. This MLP is pre-trained on 100 SGGX BRDFs with random parameters~\cite{heitz2015sggx}. As shown in the figure, Solely using a dedicated appearance MLP fails to model complex all-frequency appearance, unlike our representation.

\subsection{Shadow Splatting vs. MLP}
We train our representation without using the shadow splatting process (Sec. 4.2 in our paper), and directly adopt the refinement MLP alone to represent the shadow function. As shown in the figure, a pure MLP-based approach does not produce high-quality shadows as ours. Moreover, such an approach is prone to over-fitting (i.e., less generalizable than shadow splatting).

\subsection{Shadow Refinement} 
We train our representation using the shadow value directly computed from splatting without the refinement MLP (Sec. 4.2 in our paper). As shown in~\figref{fig:ablation}, while the main features are preserved, the shadows are not as detailed as our pipeline with the refinement step.

\subsection{MLP for Other Effects} 
We train our representation with the MLP for approximating other effects removed (Sec. 4.3 in our paper). In ~\figref{fig:ablation}, we can observe noisy results where effects like interreflections should be presented. Other components in the representation tend to compensate for the effects that would have been modeled by this missing MLP, making it more prone to over-fitting.

\subsection{Number of Basis Angular Gaussians} 
In ~\figref{fig:ab_asgnum}, we evaluate the impact of the number of basis angular Gaussians on a specific scene. And \tabref{tab:ablation} lists the quantitative scores averaged over all synthetic objects/scenes. Note that the qualitative differences are more obvious in the figure, compared with the differences in average scores in the table. Our current selection of 8 is made after balancing under- and over-fitting of appearance.

\subsection{Number of Input Images} 
\figref{fig:ab_photonum} evaluates the impact of input image number. And the average errors are also reported in~\tabref{tab:ablation}. More input images improve the reconstruction quality, due to more sampling in the view/lighting domain. We observe that it requires less input images to faithfully reconstruct scenes with simple geometry and appearance (e.g., {\sc{Hotdog}}) than more complex ones.

\begin{table}
  \caption{Ablation studies of components in our represenetation. We list the average quantitative errors in SSIM, PSNR and LPIPS of all synthetic scenes to quantify the impact of ablated components. Our choices for individual components are shown in bold.}
  \label{tab:ablation}
  \begin{tabular}{l|ccc}
    \toprule
    Ablation Variant & SSIM$\uparrow$ & PSNR$\uparrow$ & LPIPS$\downarrow$ \\
    \midrule
    \textbf{Full} & 0.9715 & 31.39 & 0.0355\\
    w/o shadow splatting & 0.9661 & 29.93 & 0.0391 \\
    w/o $\Phi$ (shadow refining) & 0.9514 & 28.03 & 0.0556 \\
    w/o $\Psi$ (other effects) & 0.9707 & 31.30 & 0.0366 \\
    \midrule
    1 basis angular Gaussians & 0.9655 & 29.70 & 0.0407 \\
    2 basis angular Gaussians & 0.9694 & 30.93 & 0.0377 \\
    4 basis angular Gaussians & 0.9709 & 31.25 & 0.0363 \\
    \textbf{8 basis angular Gaussians} & 0.9715 & 31.39 & 0.0355 \\
    16 basis angular Gaussians & 0.9721 & 31.50 & 0.0350 \\
    \midrule
    500 images & 0.9670 & 30.61 & 0.0390 \\
    1,000 images & 0.9698 & 31.09 & 0.0370 \\
    \textbf{2,000 images} & 0.9715 & 31.39 & 0.0355  \\
  \bottomrule
\end{tabular}
\end{table}

\section{Metrics}

We report detailed quantitative metrics here, due to the limited space in the main paper. First, \tabref{tab:environment} lists the errors of the comparison experiments with alternative approaches that take in environment-lit input images (corresponding to Fig. 12 in the main paper). Next, \tabref{tab:point} reports the  errors of the comparisons with~\cite{zeng2023nrhints}, the best-quality method that takes in point-lit input images (corresponding to Fig. 9 in the main paper). Finally, \tabref{tab:metrics} shows the errors for each synthetic and captured object/scene in this paper (corresponding to Fig.~5 and~11 in the main paper). Please refer to Sec. 5 for more details about the objects/scenes.

\begin{table}
  \caption{Quantitative metrics corresponding to Fig.~12 in the main paper. We list the quantitative errors in SSIM, PSNR and LPIPS.
  }
  \label{tab:environment}
  \begin{tabular}{l|c|ccc}
    \toprule
    Scene & Method & SSIM$\uparrow$ & PSNR$\uparrow$ & LPIPS$\downarrow$ \\
    \midrule
    \multirow{5}{*}{Hotdog} & GaussianShader & 0.9439 & 29.25 & 0.0600\\
    & GS-IR  & 0.9289 & 29.13 & 0.0816 \\
    & \fontsize{8}{10}\selectfont{Relightable3DGaussian} & 0.9526 & 30.22 & 0.0555 \\
    & TensoIR &  0.9590 & 31.68 & 0.0466 \\
    & Ours & \textbf{0.9712} & \textbf{36.47} & \textbf{0.0342} \\
    \midrule
    \multirow{5}{*}{Lego} & GaussianShader & 0.8964 & 24.33 & 0.0834\\
    & GS-IR  & 0.9230 & 27.66 & 0.0550 \\
    & \fontsize{8}{10}\selectfont{Relightable3DGaussian} & 0.9463 & 30.31 & 0.0446 \\
    & TensoIR & 0.9540 & 30.96 & 0.0356 \\
    & Ours & \textbf{0.9651} & \textbf{32.06} & \textbf{0.0302} \\
    \midrule
    \multirow{5}{*}{MaterialBalls} & GaussianShader & 0.8900 & 24.31 & 0.0937\\
    & GS-IR  & 0.8756 & 23.41 & 0.0969 \\
    & \fontsize{8}{10}\selectfont{Relightable3DGaussian} & 0.9045 & 25.34 & 0.0848 \\
    & TensoIR & 0.9026 & 25.05 & 0.0765  \\
    & Ours & \textbf{0.9459} & \textbf{28.01} & \textbf{0.0508} \\
  \bottomrule
\end{tabular}
\end{table}

\begin{table}
  \caption{Per-object/scene quantitative comparison with NRHints~\cite{zeng2023nrhints} (corresponding to Fig.~9 in the main paper). We list the quantitative errors in SSIM, PSNR and LPIPS.}
  \label{tab:point}
  \begin{tabular}{l|c|ccc}
    \toprule
    Scene & Method & SSIM$\uparrow$ & PSNR$\uparrow$ & LPIPS$\downarrow$ \\
    \midrule
    \multirow{2}{*}{Drums} & NRHints & \textbf{0.9745} & 29.88 & 0.0294\\
    & Ours & 0.9714 & \textbf{30.45} & \textbf{0.0276}  \\
    \midrule
    \multirow{2}{*}{FurBall} & NRHints & 0.9522 & 35.06 & 0.0777\\
    & Ours & \textbf{0.9669} & \textbf{35.34} & \textbf{0.0534}  \\
    \midrule
    \multirow{2}{*}{Lego} & NRHints & \textbf{0.9583} & 29.90 & \textbf{0.0393}\\
    & Ours & 0.9511 & \textbf{30.19} & 0.0468 \\
    \midrule
    \multirow{2}{*}{Fish} & NRHints & 0.9140 & \textbf{31.28} & 0.1137\\
    & Ours & \textbf{0.9252} & 31.13 & \textbf{0.0866} \\
    \midrule
    \multirow{2}{*}{Cluttered} & NRHints & 0.9280 & 29.61 & 0.0879\\
    & Ours & \textbf{0.9521} & \textbf{30.82} & \textbf{0.0696}  \\
    \midrule
    \multirow{2}{*}{Cat} & NRHints & 0.8560 & \textbf{27.34} & 0.1667\\
    & Ours & \textbf{0.8981} & 26.43 & \textbf{0.1381} \\
  \bottomrule
\end{tabular}
\end{table}

\begin{table}
  \caption{Per-object/scene quantitative errors of our approach for all synthetic and captured examples.}
  \label{tab:metrics}
  \begin{tabular}{l|ccc}
    \toprule
    Scence & SSIM & PSNR & LPIPS \\
    \midrule
    AnisoMetal & 0.9494 & 27.25& 0.0462\\
    Drums & 0.9714 &30.45 & 0.0276 \\
    FurBall & 0.9669 & 35.34 & 0.0534 \\
    Hotdog & 0.9733 & 32.98 & 0.0295 \\
    Lego & 0.9511 & 30.19 & 0.0468 \\
    Translucent & 0.9740 & 32.34 & 0.0318 \\
    \midrule
    Cup & 0.9905 & 32.74 & 0.0203 \\
    Egg & 0.9833 & 31.84 & 0.0257 \\
    Fabric & 0.9812 & 31.15 & 0.0317  \\
    MaterialBalls & 0.9545 & 28.04 & 0.0628 \\
    Tower & 0.9913 & 32.96 & 0.0150 \\
    \midrule
    Boot & 0.8980 & 28.84 & 0.1013 \\
    Container & 0.9745 & 36.65 & 0.0160 \\
    Fox & 0.9225 & 34.21 & 0.0745  \\
    Li'lOnes & 0.9809 & 38.61 & 0.0182  \\
    Nefertiti & 0.956 & 36.58 & 0.0434  \\
    Zhaojun & 0.9321 & 32.08 & 0.1071  \\
  \bottomrule
\end{tabular}
\end{table}

\input{figure/figure_only/ablation}

\section{Visualization}
In addition to Fig. 8 of the main paper, here we visualize of the normals of spatial Gaussians for additional scenes in~\figref{fig:vis_normal}.

\input{figure/figure_only/vis_normal}

%% file: figure/figure_only/ab_asgnum.tex
\begin{figure*}[htb]

    \begin{minipage}{\linewidth}
        \centering
        \begin{minipage}{1.15in}
            \centering
            \scalebox{.85}{Ground-Truth}
        \end{minipage}
        \begin{minipage}{1.15in}
            \centering
            \scalebox{.85}{16 Angular Gaussians}
        \end{minipage}
        \begin{minipage}{1.15in}
            \centering
            \scalebox{.85}{8 Angular Gaussians}
        \end{minipage}
        \begin{minipage}{1.15in}
            \centering
            \scalebox{.85}{4 Angular Gaussians}
        \end{minipage}
        \begin{minipage}{1.15in}
            \centering
            \scalebox{.85}{2 Angular Gaussians}
        \end{minipage}
        \begin{minipage}{1.15in}
            \centering
            \scalebox{.85}{1 Angular Gaussians}
        \end{minipage}
    \end{minipage}
    
    \begin{minipage}{\linewidth}
        \centering
        \begin{minipage}{1.15in}
            \includegraphics[width=\textwidth]{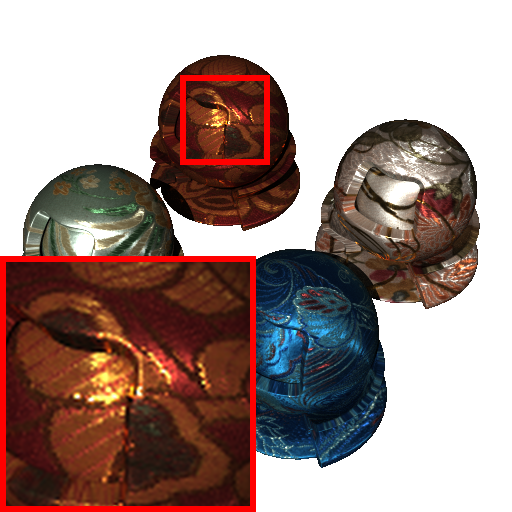}
        \end{minipage}
        \begin{minipage}{1.15in}
            \includegraphics[width=\textwidth]{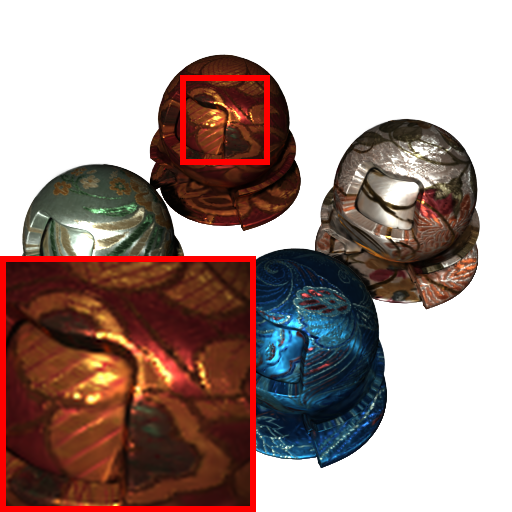}
        \end{minipage}
        \begin{minipage}{1.15in}
            \includegraphics[width=\textwidth]{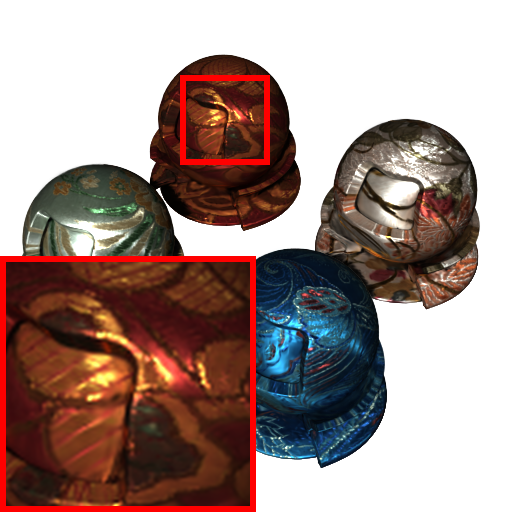}
        \end{minipage}
        \begin{minipage}{1.15in}
            \includegraphics[width=\textwidth]{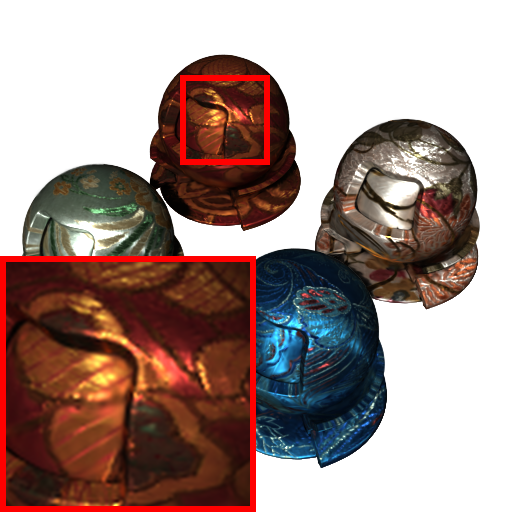}
        \end{minipage}
        \begin{minipage}{1.15in}
            \includegraphics[width=\textwidth]{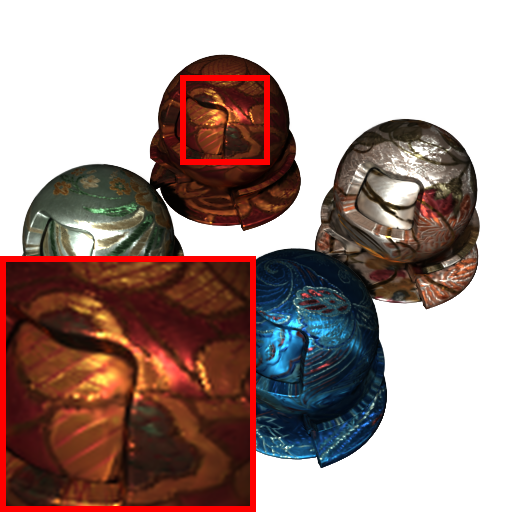}
        \end{minipage}
        \begin{minipage}{1.15in}
            \includegraphics[width=\textwidth]{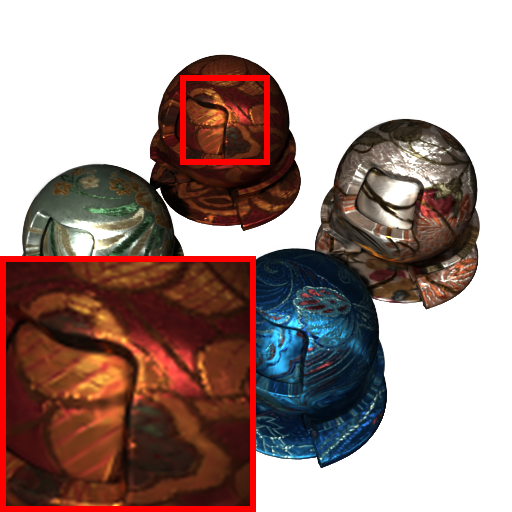}
        \end{minipage}
    \end{minipage}

    \begin{minipage}{\linewidth}
        \centering
        \begin{minipage}{1.15in}
            \centering
            \scalebox{.85}{SSIM | PSNR | LPIPS}
        \end{minipage}
        \begin{minipage}{1.15in}
            \centering
            \scalebox{.85}{0.9557 | 28.22 | 0.0618}
        \end{minipage}
        \begin{minipage}{1.15in}
            \centering
            \scalebox{.85}{0.9545 | 28.04 | 0.0628}
        \end{minipage}
        \begin{minipage}{1.15in}
            \centering
            \scalebox{.85}{0.9521 | 27.80 | 0.0649}
        \end{minipage}
        \begin{minipage}{1.15in}
            \centering
            \scalebox{.85}{0.9497 | 27.64 | 0.0674}
        \end{minipage}
        \begin{minipage}{1.15in}
            \centering
            \scalebox{.85}{0.9459 | 27.25 | 0.0710}
        \end{minipage}
    \end{minipage}

    \caption{Impact of the number of angular Gaussians. From the left image to right, the ground-truth,  results from our representations trained with different numbers of basis angular Gaussians. Average errors in SSIM, PSNR and LPIPS are reported at the bottom of each related image.}
    \label{fig:ab_asgnum}
\end{figure*}

%% file: figure/figure_only/ab_photonum.tex
\begin{figure}[htb]

    \begin{minipage}{\linewidth}
        \centering
        \begin{minipage}{0.82in}
            \centering
            \scalebox{.85}{Ground-Truth}
        \end{minipage}
        \begin{minipage}{0.82in}
            \centering
            \scalebox{.85}{500}
        \end{minipage}
        \begin{minipage}{0.82in}
            \centering
            \scalebox{.85}{1,000}
        \end{minipage}
        \begin{minipage}{0.82in}
            \centering
            \scalebox{.85}{2,000}
        \end{minipage}
    \end{minipage}
    
    \begin{minipage}{\linewidth}
        \centering
        \begin{minipage}{0.82in}
            \includegraphics[width=\textwidth]{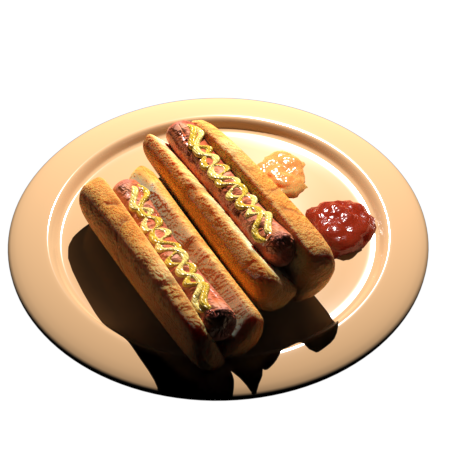}
        \end{minipage}
        \begin{minipage}{0.82in}
            \includegraphics[width=\textwidth]{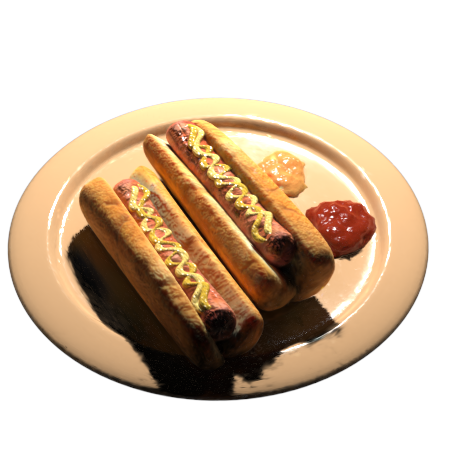}
        \end{minipage}
        \begin{minipage}{0.82in}
            \includegraphics[width=\textwidth]{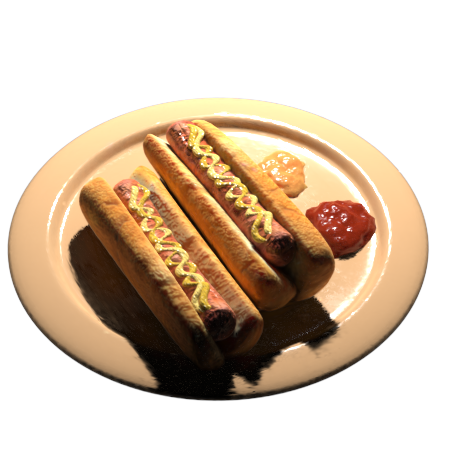}
        \end{minipage}
        \begin{minipage}{0.82in}
            \includegraphics[width=\textwidth]{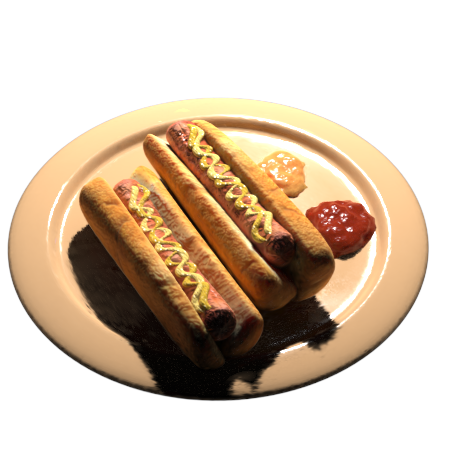}
        \end{minipage}
    \end{minipage}

    \begin{minipage}{\linewidth}
        \centering
        \begin{minipage}{0.82in}
            \centering
            \scalebox{.73}{SSIM | PSNR | LPIPS}
        \end{minipage}
        \begin{minipage}{0.82in}
            \centering
            \scalebox{.73}{0.9689 | 31.30 | 0.0327}
        \end{minipage}
        \begin{minipage}{0.82in}
            \centering
            \scalebox{.73}{0.9715 | 32.30 | 0.0308}
        \end{minipage}
        \begin{minipage}{0.82in}
            \centering
            \scalebox{.73}{0.9733 | 32.98 | 0.0295}
        \end{minipage}
    \end{minipage}
    
    \begin{minipage}{\linewidth}
        \centering
        \begin{minipage}{0.82in}
            \includegraphics[width=\textwidth]{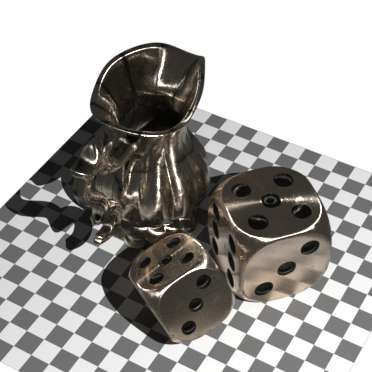}
        \end{minipage}
        \begin{minipage}{0.82in}
            \includegraphics[width=\textwidth]{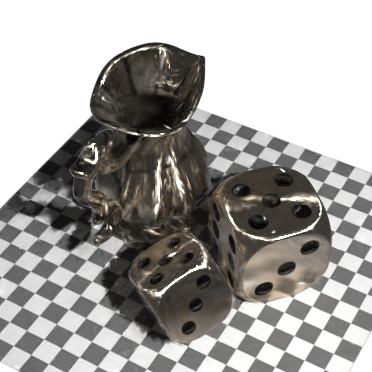}
        \end{minipage}
        \begin{minipage}{0.82in}
            \includegraphics[width=\textwidth]{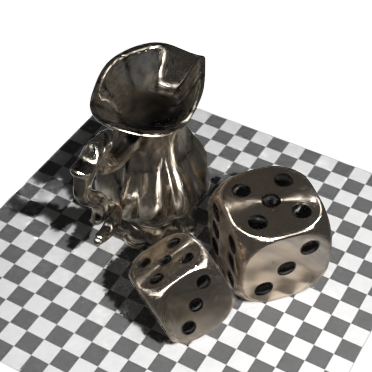}
        \end{minipage}
        \begin{minipage}{0.82in}
            \includegraphics[width=\textwidth]{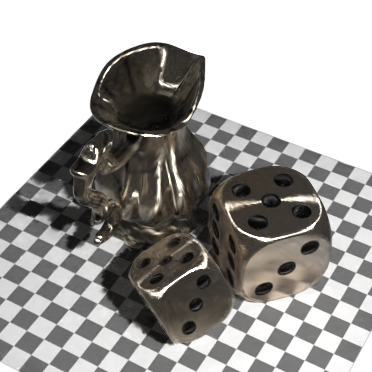}
        \end{minipage}

        \begin{minipage}{\linewidth}
        \centering
        \begin{minipage}{0.82in}
            \centering
            \scalebox{.73}{SSIM | PSNR | LPIPS}
        \end{minipage}
        \begin{minipage}{0.82in}
            \centering
            \scalebox{.73}{0.9383 | 26.20 | 0.0530}
        \end{minipage}
        \begin{minipage}{0.82in}
            \centering
            \scalebox{.73}{0.9457 | 26.95 | 0.0483}
        \end{minipage}
        \begin{minipage}{0.82in}
            \centering
            \scalebox{.73}{0.9494 | 27.25 | 0.0462}
        \end{minipage}
    \end{minipage}
    
    \end{minipage}
    \caption{Impact of the number of training images. From the left column to right, the ground-truths, and reconstruction results trained with 500, 1,000 and 2,000 images. Average errors in SSIM, PSNR and LPIPS are reported at the bottom of each related image.}
    \label{fig:ab_photonum}
\end{figure}

%% file: figure/figure_only/ablation.tex
\begin{figure}[htb]

    \begin{minipage}{\linewidth}
        \centering
        \begin{minipage}{0.06in}
        \end{minipage}
        \begin{minipage}{1.0in}
            \centering
            \scalebox{.85}{Ground-Truth}
        \end{minipage}
        \begin{minipage}{1.0in}
            \centering
            \scalebox{.85}{Ours}
        \end{minipage}
        \begin{minipage}{1.0in}
            \centering
            \scalebox{.85}{Ablation Variant}
        \end{minipage}
    \end{minipage}

    \begin{minipage}{\linewidth}
        \centering
        \begin{minipage}{0.06in}
            \rotatebox{90}{\scalebox{.85}{Naive MLP}}
        \end{minipage}
        \begin{minipage}{1.0in}
            \includegraphics[width=\textwidth]{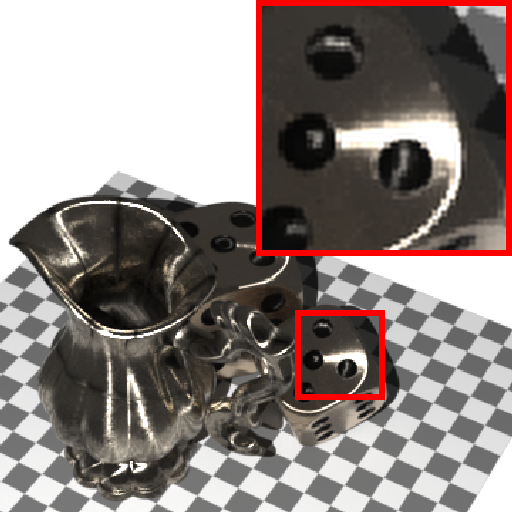}
        \end{minipage}
        \begin{minipage}{1.0in}
            \includegraphics[width=\textwidth]{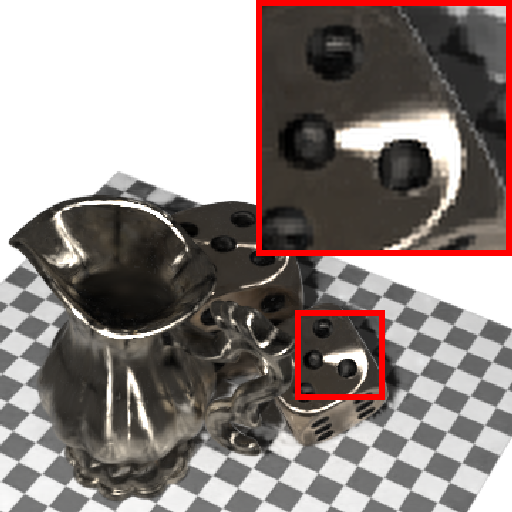}
        \end{minipage}
        \begin{minipage}{1.0in}
            \includegraphics[width=\textwidth]{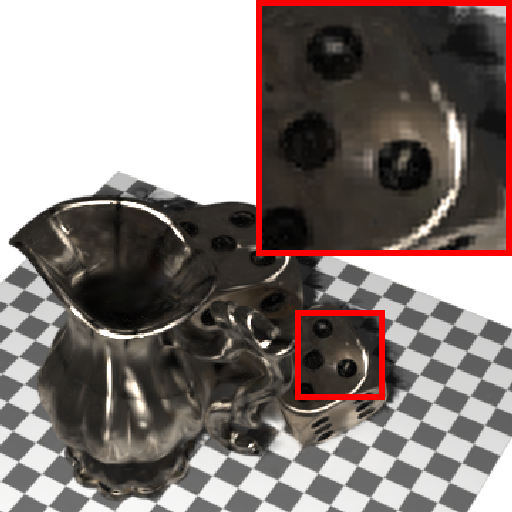}
        \end{minipage}
    \end{minipage}

    \begin{minipage}{\linewidth}
        \centering
        \begin{minipage}{0.06in}
        \end{minipage}
        \begin{minipage}{1.0in}
            \centering
            \scalebox{.85}{\sc{SSIM | PSNR | LPIPS}}
        \end{minipage}
        \begin{minipage}{1.0in}
            \centering
            \scalebox{.85}{\sc{0.9494 | 27.25 | 0.0462 }}
        \end{minipage}
        \begin{minipage}{1.0in}
            \centering
            \scalebox{.85}{\sc{0.9469 | 26.29 | 0.0479}}
        \end{minipage}
    \end{minipage}

    \begin{minipage}{\linewidth}
        \centering
        \begin{minipage}{0.06in}
            \rotatebox{90}{\scalebox{.85}{w/o $\Phi$ (Shadow Refinement)}}
        \end{minipage}
        \begin{minipage}{1.0in}
            \includegraphics[width=\textwidth]{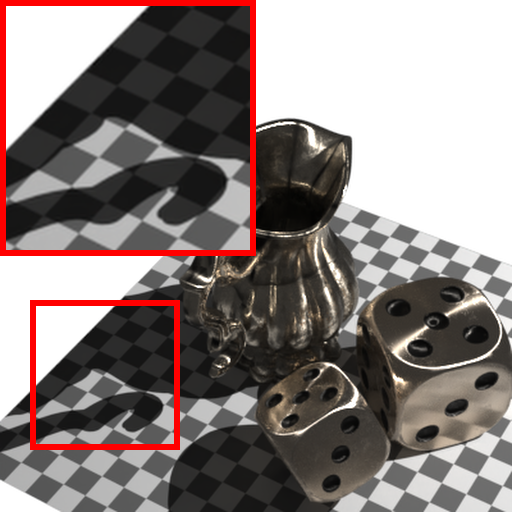}
        \end{minipage}
        \begin{minipage}{1.0in}
            \includegraphics[width=\textwidth]{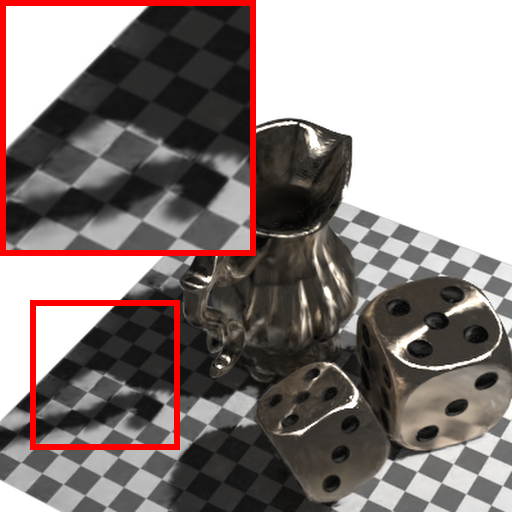}
        \end{minipage}
        \begin{minipage}{1.0in}
            \includegraphics[width=\textwidth]{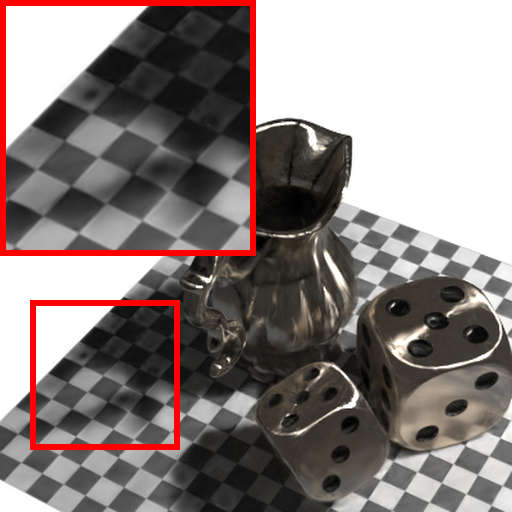}
        \end{minipage}
    \end{minipage}

    \begin{minipage}{\linewidth}
        \centering
        \begin{minipage}{0.06in}
        \end{minipage}
        \begin{minipage}{1.0in}
            \centering
            \scalebox{.85}{\sc{SSIM | PSNR | LPIPS}}
        \end{minipage}
        \begin{minipage}{1.0in}
            \centering
            \scalebox{.85}{\sc{0.9494 | 27.25 | 0.0462 }}
        \end{minipage}
        \begin{minipage}{1.0in}
            \centering
            \scalebox{.85}{\sc{0.9359 | 25.74 | 0.0552}}
        \end{minipage}
    \end{minipage}

    \begin{minipage}{\linewidth}
        \centering
        \begin{minipage}{0.06in}
            \rotatebox{90}{\scalebox{.85}{w/o $\Psi$ (Other Effects)}}
        \end{minipage}
        \begin{minipage}{1.0in}
            \includegraphics[width=\textwidth]{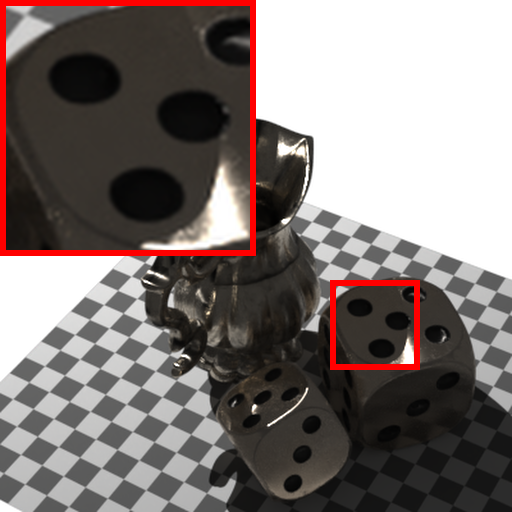}
        \end{minipage}
        \begin{minipage}{1.0in}
            \includegraphics[width=\textwidth]{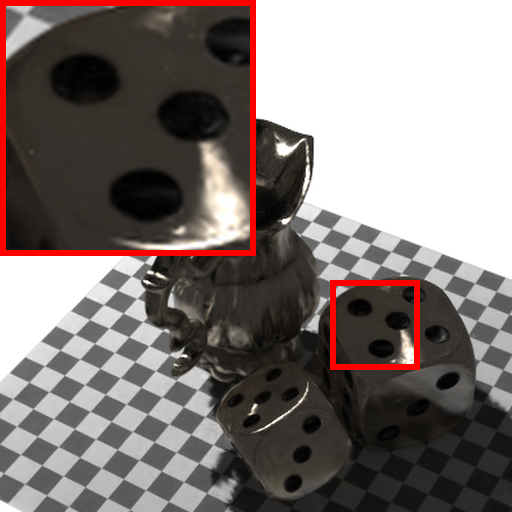}
        \end{minipage}
        \begin{minipage}{1.0in}
            \includegraphics[width=\textwidth]{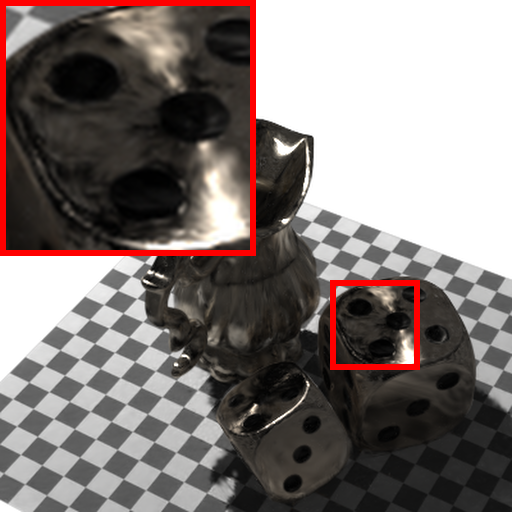}
        \end{minipage}
    \end{minipage}

    \begin{minipage}{\linewidth}
        \centering
        \begin{minipage}{0.06in}
        \end{minipage}
        \begin{minipage}{1.0in}
            \centering
            \scalebox{.85}{\sc{SSIM | PSNR | LPIPS}}
        \end{minipage}
        \begin{minipage}{1.0in}
            \centering
            \scalebox{.85}{\sc{0.9494 | 27.25 | 0.0462 }}
        \end{minipage}
        \begin{minipage}{1.0in}
            \centering
            \scalebox{.85}{\sc{0.9450 | 26.92 | 0.0490}}
        \end{minipage}
    \end{minipage}

    \begin{minipage}{\linewidth}
        \centering
        \begin{minipage}{0.06in}
            \rotatebox{90}{\scalebox{.85}{w/o Shadow Splatting}}
        \end{minipage}
        \begin{minipage}{1.0in}
            \includegraphics[width=\textwidth]{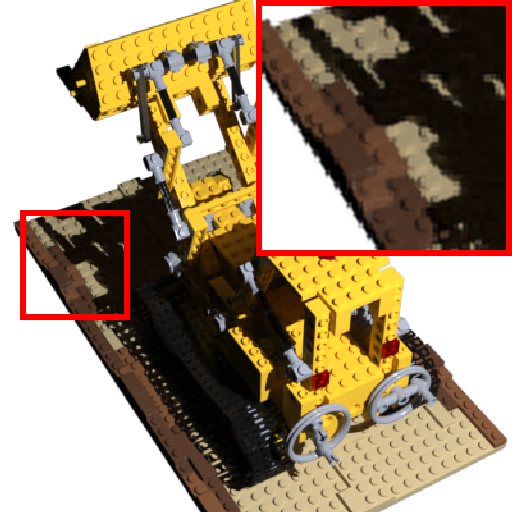}
        \end{minipage}
        \begin{minipage}{1.0in}
            \includegraphics[width=\textwidth]{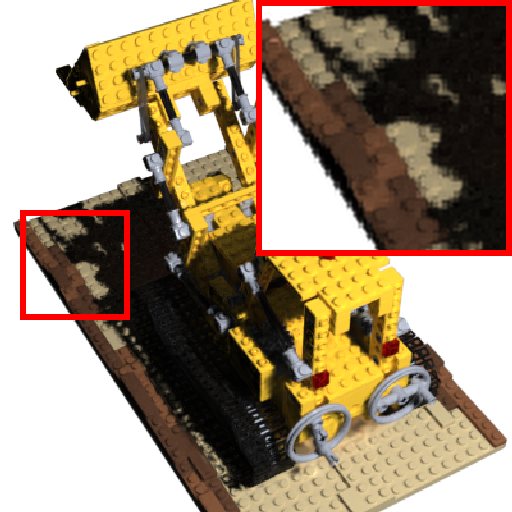}
        \end{minipage}
        \begin{minipage}{1.0in}
            \includegraphics[width=\textwidth]{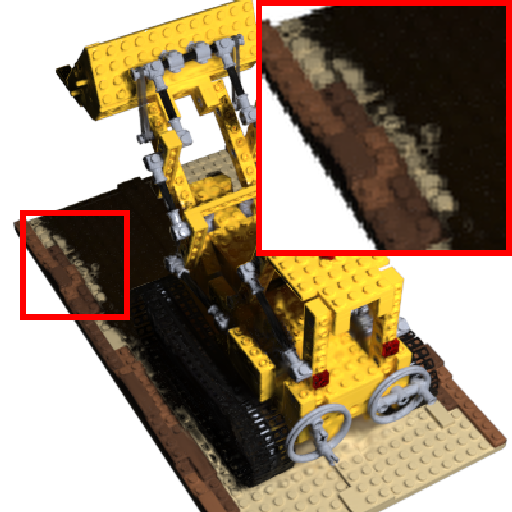}
        \end{minipage}
    \end{minipage}

    \begin{minipage}{\linewidth}
        \centering
        \begin{minipage}{0.06in}
        \end{minipage}
        \begin{minipage}{1.0in}
            \centering
            \scalebox{.85}{\sc{SSIM | PSNR | LPIPS}}
        \end{minipage}
        \begin{minipage}{1.0in}
            \centering
            \scalebox{.85}{\sc{0.9511 | 30.19 | 0.0468 }}
        \end{minipage}
        \begin{minipage}{1.0in}
            \centering
            \scalebox{.85}{\sc{0.9272 | 26.03 | 0.0637}}
        \end{minipage}
    \end{minipage}

    \caption{Ablation studies on components of our pipeline. From the left column to right, the ground-truth, our results and results from the variants. From the top row to bottom, solely using an MLP as appearance function, removing shadow refinement step, removing the modeling of other effects, and removing shadow splatting and directly resorting to an MLP. Average errors in SSIM, PSNR and LPIPS are reported at the bottom of each related image.}
    
    \label{fig:ablation}
\end{figure}

%% file: figure/figure_only/vis_normal.tex
\begin{figure}[htb]

    \begin{minipage}{\linewidth}
        \centering
        \begin{minipage}{0.82in}
            \centering
            \scalebox{.85}{\sc{Cup}}
        \end{minipage}
        \begin{minipage}{0.82in}
            \centering
            \scalebox{.85}{\sc{Hotdog}}
        \end{minipage}
        \begin{minipage}{0.82in}
            \centering
            \scalebox{.85}{\sc{Lego}}
        \end{minipage}
        \begin{minipage}{0.82in}
            \centering
            \scalebox{.85}{\sc{Furball}}
        \end{minipage}
    \end{minipage}
    
    \begin{minipage}{\linewidth}
        \centering
        \begin{minipage}{0.82in}
            \includegraphics[width=\textwidth]{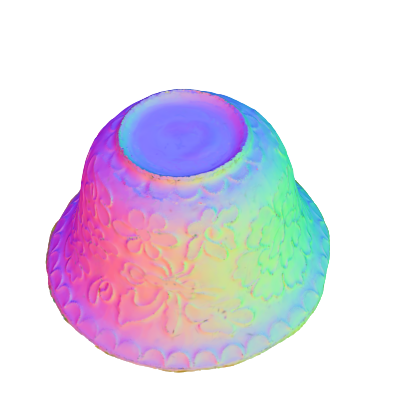}
        \end{minipage}
        \begin{minipage}{0.82in}
            \includegraphics[width=\textwidth]{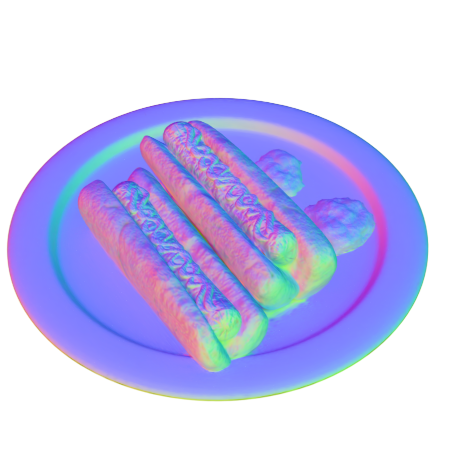}
        \end{minipage}
        \begin{minipage}{0.82in}
            \includegraphics[width=\textwidth]{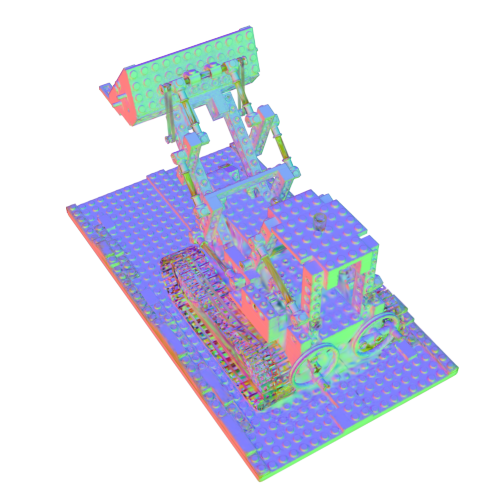}
        \end{minipage}
        \begin{minipage}{0.82in}
            \includegraphics[width=\textwidth]{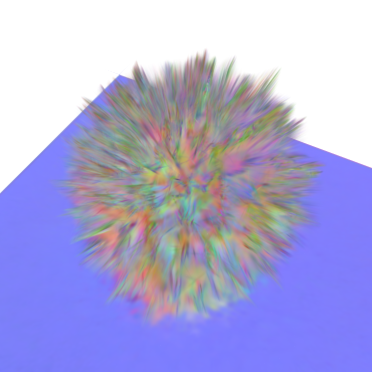}
        \end{minipage}
    \end{minipage}
    
    \caption{Visualization of the normals of spatial Gaussians in additional scenes. Each spatial Gaussian is splatted with the pseudo color of its normal.}
    \label{fig:vis_normal}
\end{figure}